\documentclass{article} 
\usepackage{iclr2027_conference,times}

\usepackage{amsmath,amsfonts,bm}

\def\eqref#1{equation~\ref{#1}}

\def\1{\bm{1}}

\def\ra{{\textnormal{a}}}

\def\rx{{\textnormal{x}}}

\def\rva{{\mathbf{a}}}

\def\erva{{\textnormal{a}}}

\def\ervx{{\textnormal{x}}}

\def\rmA{{\mathbf{A}}}

\def\vmu{{\bm{\mu}}}
\def\vtheta{{\bm{\theta}}}
\def\va{{\bm{a}}}

\def\ve{{\bm{e}}}

\def\vx{{\bm{x}}}

\def\eva{{a}}

\def\mA{{\bm{A}}}

\def\mH{{\bm{H}}}
\def\mI{{\bm{I}}}
\def\mJ{{\bm{J}}}

\def\mX{{\bm{X}}}

\def\mSigma{{\bm{\Sigma}}}

\DeclareMathAlphabet{\mathsfit}{\encodingdefault}{\sfdefault}{m}{sl}
\SetMathAlphabet{\mathsfit}{bold}{\encodingdefault}{\sfdefault}{bx}{n}
\newcommand{\tens}[1]{\bm{\mathsfit{#1}}}
\def\tA{{\tens{A}}}

\def\tX{{\tens{X}}}

\def\gG{{\mathcal{G}}}

\def\sA{{\mathbb{A}}}
\def\sB{{\mathbb{B}}}

\def\sS{{\mathbb{S}}}

\def\emA{{A}}

\newcommand{\etens}[1]{\mathsfit{#1}}

\def\etA{{\etens{A}}}

\newcommand{\E}{\mathbb{E}}

\newcommand{\R}{\mathbb{R}}

\newcommand{\KL}{D_{\mathrm{KL}}}
\newcommand{\Var}{\mathrm{Var}}

\newcommand{\Cov}{\mathrm{Cov}}

\newcommand{\normltwo}{L^2}
\newcommand{\normlp}{L^p}

\newcommand{\parents}{Pa} 

\usepackage{graphicx}
\usepackage{subcaption}
\usepackage{booktabs}
\usepackage{enumitem}
\usepackage{tabularx}
\usepackage{multirow}
\usepackage{amssymb}
\usepackage[autostyle=true]{csquotes}

\usepackage{needspace}

\usepackage{amsthm}
\theoremstyle{plain}

\usepackage{hyperref}
\usepackage{url}

\title{Augmenting Visual Anomaly Detection with Automated Interpretability}

\author{\textbf{Antonio De Santis }\normalfont{\textsuperscript{1,2}}\thanks{Work done while interning at Thales Alenia Space. Email: \texttt{antonio.desantis@polimi.it}.}\quad\quad
\textbf{Arsenio Leo}\textsuperscript{2}\quad\quad
\textbf{Marco Brambilla}\textsuperscript{1} \\
 \\
\textsuperscript{1}Politecnico di Milano \quad \textsuperscript{2}Thales Alenia Space
}

\iclrfinalcopy 
\begin{document}

\maketitle
\lhead{}
\renewcommand{\headrulewidth}{0pt}

\begin{abstract}
Visual anomaly detectors identify deviations from known-normal data, but their
anomaly signals may mix evidence of actual anomalies with benign visual
variation. We investigate whether automated interpretability can augment visual
anomaly detectors by identifying and intervening on different components of
this signal. We decompose PatchCore nearest-normal residuals into sparse
features using Sparse Autoencoders (SAEs), and provide high-activation and
contrastive non-active examples to a Multimodal LLM, which describes each
feature and labels it as anomaly, distractor, or uncertain. These labels guide
interventions in the SAE hidden representation, where distractor features are
suppressed and anomaly features amplified. The edited
representation is then used to reconstruct patch embeddings, which are
rescored with PatchCore. Across 40 categories from four benchmarks, applying
both interventions jointly improves macro-average image-level AUROC from 0.8724
to 0.8857 on source data and from 0.8066 to 0.8210 under synthetic corruptions.
On three additional RobustAD categories with real acquisition shifts, the same
interventions improve AUROC from 0.8745 to 0.9056 on source data and from
0.6069 to 0.6599 under real acquisition shifts. Finally, individual feature interventions across all 43 categories show that the MLLM labels are aligned in
aggregate with how features differently affect normal and anomalous
images.
\end{abstract}

\section{Introduction}
\label{intro}
Visual anomaly detection aims to identify anomalies without requiring examples
of all possible anomaly types. Methods such as PatchCore \citep{patchcore}
approach this problem by comparing representations of an input image with
representations extracted from known-normal data. However, not every difference
from known-normal examples in embedding space reflects a defect or a violation
of the expected object configuration. Benign visual changes, such as differences
in background, lighting, viewpoint, surface appearance, or manufacturing
variations that still fall within the definition of normality, can also move an
input representation away from its normal reference.
The resulting signal may therefore mix evidence of actual anomalies with
variation that should ideally not affect the detector's decision. Recent work
in mechanistic interpretability has used Sparse Autoencoders (SAEs) to
decompose neural representations into sparse features
\citep{bricken2023monosemanticity,gao2025scaling}, and language or multimodal
models to generate human-understandable descriptions of these features
\citep{templeton2024scaling,desantis2026learning}. This motivates the question of
whether these automated interpretations can separate task-relevant anomaly evidence
from distracting variation well enough to guide interventions on the detector
itself.

In this paper, we investigate this question by decomposing the anomaly
residuals of a frozen visual anomaly detector into sparse features and
automatically interpreting what these features represent. Using a small calibration
set containing both normal and anomalous images, we compute residuals between
patch embeddings and their nearest known-normal embeddings and
decompose these residuals with an SAE. We then provide high-activation and contrastive
examples of each learned feature to a Multimodal Large Language Model (MLLM),
which describes the visual pattern associated with the feature and labels it
as \emph{anomaly}, \emph{distractor}, or \emph{uncertain}. These labels are used to suppress distractor features and amplify anomaly
features. The resulting patch embeddings are then rescored by the same frozen
anomaly detector. Features labeled as \emph{uncertain} are left unchanged.

We organize our study around three main research questions. \textbf{RQ1} asks
whether feature interventions guided by automated interpretability improve
anomaly detection performance on source data. \textbf{RQ2} asks whether these
interventions remain effective under synthetic corruptions and real acquisition
shifts. \textbf{RQ3} asks to what extent the labels assigned by the MLLM
correspond to the effects of suppressing or amplifying individual SAE features
in isolation. Our results show that the proposed feature interventions improve PatchCore
performance on source data across 40 categories from four visual anomaly
detection benchmarks. These gains persist under synthetic corruptions. We also
evaluate three additional categories from RobustAD, a benchmark with real
acquisition shifts, where the interventions produce the largest improvements.
Finally, individual feature interventions across all 43 categories show that
the MLLM labels are aligned in aggregate with how the corresponding features differently
affect normal and anomalous images.

\section{Related Work}
\label{related_work}

\paragraph{Mechanistic Interpretability.}
Mechanistic interpretability aims to reverse-engineer neural networks by
understanding the roles of their internal components, such as neurons and
their interactions. Directly interpreting neurons is challenging because they often do not map one-to-one to human-understandable concepts. A single neuron can respond to multiple unrelated concepts
\citep{netdissect2017}, a phenomenon known as polysemanticity
\citep{olah2020zoom}. A prominent explanation for polysemanticity is
superposition, where a network represents more features than there are
available activation dimensions by encoding them along non-orthogonal
directions \citep{elhage2022toymodelssuperposition}. One way to avoid relying on individual neurons is to represent concepts as
directions in activation space, which can be constructed from user-provided
examples \citep{kim2018interpretability,desantis2026visualtcav}. Related work
has shown that directions in activation space can also be used to steer model
behavior \citep{rimsky-etal-2024-steering}. Concept discovery methods can instead identify these concept directions
directly by applying clustering or factorization techniques in the model activation space
\citep{ace,ice,craft,abc}. Sparse Autoencoders (SAEs) extend this idea by learning a sparse dictionary of
features from model activations with an autoencoder, with only a small number
of features active for each input, providing a way to recover features
represented in superposition \citep{bricken2023monosemanticity}. 

Automated interpretability refers to using language or multimodal models to automatically map these discovered features to human-understandable concepts. Early
work used language models to describe neurons from their own activation examples
\citep{bills2023language}, while subsequent work extended this approach to
vision models \citep{shaham2024multimodal}. \cite{desantis2026learning} further combined SAEs with MLLMs to extract and automatically label concepts learned by vision backbones. Sparse features can also provide an
interface for editing internal representations
\citep{bricken2023monosemanticity,marks2025sparse}. In our
setting, we use an MLLM to label sparse SAE features that decompose
the anomaly signal of a visual anomaly detector, and use these labels to automatically determine which features should be suppressed or amplified.

\paragraph{Visual Anomaly Detection.}

Visual anomaly detection typically models normal data and identifies inputs that
deviate from it, avoiding the assumption that all relevant anomaly types are observed during training. Existing approaches model normality in different ways, including feature distributions \citep{padim,cflow}, stored normal representations \citep{patchcore}, and discrepancies between learned representations \citep{reverse_distillation,efficientad,fre}. PatchCore \citep{patchcore}, which we use in our main experiments, stores normal patch embeddings in a memory bank and scores test images using their distance to these embeddings. FRE \citep{fre}, which we use as an alternative detector in Appendix~\ref{app:detector-ablation}, instead models normality using a shallow linear autoencoder and scores anomalies from its reconstruction error.
In terms of interpretability, prior work has
primarily focused on local explanations and spatial localization of anomalous
regions \citep{wang2025unveiling,liznerski2021explainable,
jiang2023interpretability}. In contrast, we decompose the anomaly signal into a
global dictionary of features, assign them semantic interpretations, and use these
interpretations to guide interventions on the detector.

One prominent approach to evaluating robustness in these visual anomaly detectors is to apply synthetic image corruptions at multiple severity levels \citep{hendrycks2018benchmarking,mvtec-c}. Such benchmarks apply controlled perturbations to the source images, such as noise, blur, and contrast variations. \citet{robustad} later introduced RobustAD, a benchmark to instead investigate robustness  in real-world scenarios. To do so, they collected real images under controlled variations in lighting, object pose, and background, and showed that synthetic corruptions do not fully reflect robustness to real acquisition changes. For this reason, we evaluate our interventions under both synthetic corruptions and the real acquisition shifts of RobustAD.

\begin{figure}[t]
    \centering
    \includegraphics[width=1\linewidth]{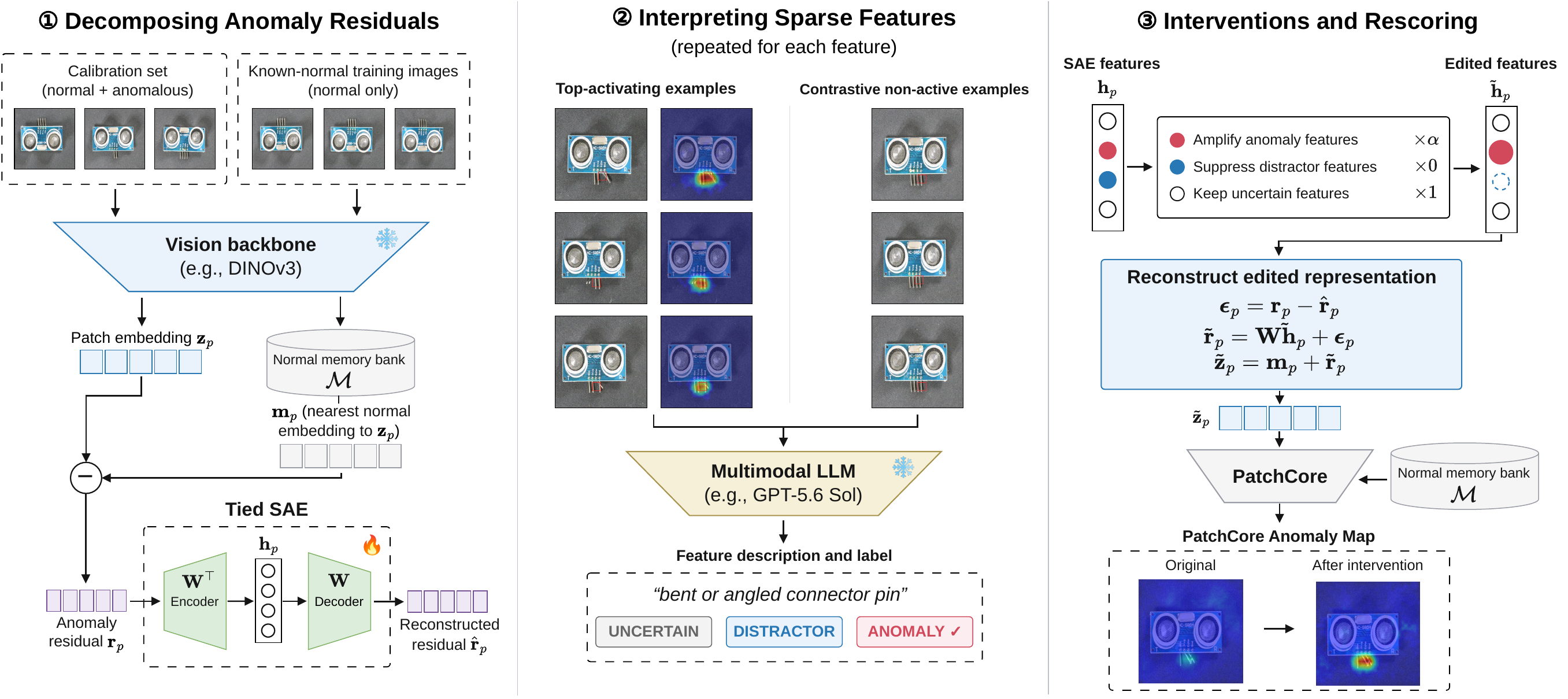}
    \caption{
    Overview of the pipeline. (1) Given a frozen vision backbone and a PatchCore
    memory bank, we compute anomaly residuals relative to the nearest normal patch
    embeddings and decompose them using a tied Sparse Autoencoder (SAE).
    (2) A multimodal LLM is prompted with top-activating and contrastive non-active
    examples to generate SAE feature descriptions and assign an
    anomaly, distractor, or uncertain label.
    (3) We amplify anomaly features and suppress distractor ones. Then we reconstruct the edited patch embeddings and rescore
    them with PatchCore.
    }
    \label{fig:method}
\end{figure}

\section{Methodology}
\label{methodology}
Our methodology consists of three steps illustrated in Figure~\ref{fig:method}. We decompose deviations from normality into sparse features using a Sparse Autoencoder (SAE), interpret these features using a Multimodal Large Language Model (MLLM), and intervene on the interpreted features before rescoring the edited representations with the same frozen anomaly detector.

\paragraph{Decomposing Anomaly Residuals.}
\label{sec:residual-decomposition}

Our starting point is a frozen visual anomaly detector that compares the representations of an input image against a known representation of normality. In our main experiments, we instantiate this setting using PatchCore \citep{patchcore}. Given an input image
$\vx$, a frozen backbone $\phi$ produces a set of patch embeddings
$\mathbf{z}_p \in \R^d$, while PatchCore stores embeddings extracted from
known-normal training images in a dataset-specific memory bank $\mathcal{M}$.
For each patch $p$, PatchCore retrieves its nearest normal embedding:
\begin{equation}
\label{eq:nearest-normal}
\mathbf{m}_p
=
\underset{\mathbf{m}\in\mathcal{M}}{\arg\min}
\;
\left\|
\mathbf{z}_p-\mathbf{m}
\right\|_2
\end{equation}
The nearest-neighbor distance
$\|\mathbf{z}_p-\mathbf{m}_p\|_2$ provides the magnitude of the patch-level anomaly evidence that is subsequently aggregated by the PatchCore scoring rule.
We define the associated deviation in the embedding space as the following residual term:
\begin{equation}
\label{eq:residual}
\mathbf{r}_p
=
\mathbf{z}_p-\mathbf{m}_p
\in\R^d
\end{equation}
We train a Sparse Autoencoder (SAE) on these patch residuals computed from a balanced calibration set containing normal and anomalous images, with all calibration
residuals defined relative to the same normal memory bank $\mathcal{M}$, as shown in
step~\textcircled{1} of Figure~\ref{fig:method}. Although our main experiments use PatchCore, the method only requires an anomaly detector that defines a deviation in embedding space from a normal reference. We
ablate the choice of anomaly detector in Appendix~\ref{app:detector-ablation}
by replacing PatchCore with FRE \citep{fre}.

The SAE uses TopK sparsity, which directly controls the number of active
features and has been shown to provide a Pareto improvement over $L_1$ regularization in the sparsity and reconstruction error trade-off
\citep{gao2025scaling}. We additionally tie encoder and decoder weights and do not use biases.
Following \cite{bricken2023monosemanticity}, we also constrain the dictionary
vectors to unit $L_2$ norm. Together with the ReLU, these constraints
ensure that, all else equal, an active feature makes the residual larger in
magnitude than it would be if that feature were inactive. We do this to encourage a dictionary of features that account for deviations
from normality, rather than encoding normality itself. In
Appendix~\ref{app:sae}, we compare this configuration with an
untied TopK SAE with learned biases and observe comparable reconstruction error
and recovered anomaly detection performance.

Let $\mathbf{W}\in\R^{d\times m}$ denote the shared encoder-decoder weight
matrix. For patch $p$, the sparse hidden representation and reconstructed
residual are:
\begin{equation}
\label{eq:sae}
\mathbf{h}_p
=
\operatorname{TopK}_{k}
\left(
\operatorname{ReLU}
\left(
\mathbf{W}^{\top}\mathbf{r}_p
\right)
\right),
\qquad
\hat{\mathbf{r}}_p
=
\mathbf{W}\mathbf{h}_p
\end{equation}
where $\operatorname{TopK}_{k}$ retains the $k$ largest activations and sets all
remaining entries to zero.

We minimize mean squared reconstruction error over the calibration residuals.
Let $\mathcal{R}$ denote the collection of calibration residuals. Since
sparsity is imposed directly by the TopK operator, no additional sparsity
penalty is used:
\begin{equation}
\label{eq:sae-loss}
\mathcal{L}_{\mathrm{SAE}}
=
\frac{1}{|\mathcal{R}|d}
\sum_{\mathbf{r}\in\mathcal{R}}
\left\|
\mathbf{r}
-
\hat{\mathbf{r}}
\right\|_2^2
\end{equation}
\paragraph{Interpreting Sparse Features.}
We next interpret the sparse features learned by the SAE using a Multimodal
Large Language Model (MLLM), as illustrated in step~\textcircled{2} of
Figure~\ref{fig:method}. For each feature, we select calibration patches with
the highest activations, taking at most one patch from each image. For MVTec
AD 2, this constraint is applied at the scene level. For each selected patch,
the MLLM receives the full image with the corresponding PatchCore region
highlighted by a red box, a spatial SAE feature activation map, and a close-up
of the selected region and its surroundings. Activation maps for the
same feature use a common scale, normalized by that feature's maximum activation
over the calibration set.
We additionally provide contrastive examples of patches where the feature has
zero activation. To emphasize what distinguishes feature activation from
non-activation, we select zero-activation patches whose patch embeddings
$\mathbf{z}_p$ are most similar to the mean embedding of the selected
high-activation patches, using cosine similarity. For zero-activation examples, we likewise select at most one patch from each
image, or from each scene for MVTec AD 2. Each contrastive example shows the full
image with the selected region highlighted and a close-up of the region. The MLLM also receives known-normal images for the corresponding
category, arranged in a $3\times3$ grid, together with the dataset and category
context. For MVTec LOCO AD and RobustAD, we additionally provide the benchmark description
of the expected normal configuration.

Given this evidence, the MLLM produces a concise description of the visual
pattern associated with the feature and assigns one of three labels. It is
instructed to use \emph{anomaly} when the feature indicates an anomaly of
the inspected object or its expected configuration, \emph{distractor} when it
captures nuisance variation, visual artifacts, or ordinary variation in the
appearance of the object or scene, and \emph{uncertain} when the evidence is
mixed or unclear. If either the high-activation or zero-activation group contains fewer than two
distinct examples after the one-per-image (or one-per-scene) selection, the
feature is assigned \emph{uncertain} without querying the MLLM. An example of a complete prompt is
provided in Appendix~\ref{app:autointerp-prompt}.

\paragraph{Interventions and Rescoring.}
Finally, we use the labels assigned to the SAE features to define the interventions, as illustrated in step~\textcircled{3} of
Figure~\ref{fig:method}. We consider three interventions: distractor
suppression, anomaly amplification, and their combination, which we refer to as
joint intervention. For distractor suppression, we zero-ablate features labeled
as \emph{distractor} by setting their activations to zero, while leaving other features unchanged. For anomaly amplification, we multiply the
activations of features labeled as \emph{anomaly} by a factor
$\alpha \geq 1$, while leaving the remaining features unchanged. The joint
intervention applies both operations simultaneously. Thus, for the joint
intervention,
\begin{equation}
\label{eq:intervention}
\tilde{h}_{p,j}
=
\begin{cases}
0, & j \in \mathcal{J}_{\mathrm{dist}},\\
\alpha h_{p,j}, & j \in \mathcal{J}_{\mathrm{anom}},\\
h_{p,j}, & \text{otherwise},
\end{cases}
\end{equation}
where $\mathcal{J}_{\mathrm{dist}}$ and $\mathcal{J}_{\mathrm{anom}}$ denote
the sets of features labeled as distractors and anomalies, respectively.
Features labeled as \emph{uncertain} are therefore left unchanged. These
interventions are applied at every patch where the corresponding feature is
active.

Because the SAE does not perfectly reconstruct the original residual, we
preserve its reconstruction error during the intervention. For each patch, we
define:
\begin{equation}
\label{eq:reconstruction-error}
\boldsymbol{\epsilon}_p
=
\mathbf{r}_p-\hat{\mathbf{r}}_p
\end{equation}
and reconstruct the intervened residual and patch embedding as:
\begin{equation}
\label{eq:edited-representation}
\tilde{\mathbf{r}}_p
=
\mathbf{W}\tilde{\mathbf{h}}_p+\boldsymbol{\epsilon}_p,
\qquad
\tilde{\mathbf{z}}_p
=
\mathbf{m}_p+\tilde{\mathbf{r}}_p
\end{equation}
so that the intervention changes only the contribution represented by the SAE
features while leaving the reconstruction error unchanged. This preserves
residual information not captured by the SAE, including information that may
be relevant to anomalies not represented in the calibration set.

For interventions involving amplification, we select a single category-level
factor $\alpha$ from the fixed set $\{1,1.5,2,3,4,5\}$ that maximizes
image-level AUROC on the source calibration set. When multiple values achieve
the same maximum AUROC, we select the smallest $\alpha$. We compare this
selection procedure with fixed amplification factors in
Appendix~\ref{app:multiplier-sensitivity}. The selected value is then fixed for both source and corruption evaluations. After reconstructing the edited
patch embeddings, we perform a fresh nearest-neighbor search over the unchanged
normal memory bank $\mathcal{M}$:
\begin{equation}
\label{eq:edited-nearest-normal}
\tilde{\mathbf{m}}_p
=
\underset{\mathbf{m}\in\mathcal{M}}{\arg\min}
\;
\left\|
\tilde{\mathbf{z}}_p-\mathbf{m}
\right\|_2
\end{equation}
The resulting nearest-neighbor distances and matches are then passed through
the same frozen PatchCore scoring rule to obtain the intervened anomaly scores.

\section{Experimental Setup}
\label{sec:experimental-setup}

\paragraph{Datasets.}
We evaluate on VisA~\citep{zou2022spot}, MVTec AD~\citep{bergmann2019mvtec},
MVTec LOCO AD~\citep{bergmann2022beyond}, and
MVTec AD 2~\citep{hecklerkram2026mvtecad2}, covering 40 categories in total,
and separately on RobustAD~\citep{robustad} under real acquisition
shifts.
For each category in the first four benchmarks, we construct a balanced
calibration set, with the number of images per class corresponding to 20\% of
the smaller test class, and use the remaining images for evaluation.
The resulting calibration sets contain a median of 20 images per category.
Because RobustAD already includes anomalous images in its training split, we
instead construct the calibration set directly from the training
data, using all anomalous images and an equal number of normal images, with a
median of 150 images per category.
For MVTec AD 2, where the same physical scene is acquired under multiple
conditions, the same sampling rule is applied to scene groups, keeping all
acquisitions of the same scene together to avoid closely related views crossing
the calibration and test splits.
For MVTec LOCO AD and RobustAD, which include descriptions of the
expected normal configuration, we additionally provide these descriptions to
the MLLM.

\paragraph{Implementation Details.}
We use PatchCore with a frozen DINOv3 ViT-L/16 backbone, extracting
representations from \texttt{blocks.11} and \texttt{blocks.23} at an input
resolution of $512\times512$, with the default coreset ratio of $0.1$ and 9
neighbors for final image scoring. For each category, we train a separate SAE
with 64 features and TopK $k=4$.
SAE-specific hyperparameters and metrics, together with alternative widths and
sparsity levels, are reported in Appendix~\ref{app:sae}, and alternative
vision backbones in Appendix~\ref{app:backbone-ablation}.
For the automated interpretability step, we use GPT-5.6 Sol with reasoning effort set to
\texttt{none} and temperature $0$.
At the time of writing the paper, a complete 64-feature annotation pass costs approximately
USD 3. We evaluate open-weight MLLMs in Appendix~\ref{app:mllm-ablation}, and
provide an example annotation prompt in
Appendix~\ref{app:autointerp-prompt}.
\begin{table}[t]
\caption{Image-level AUROC on source and synthetically corrupted data, macro-averaged
across categories. Best results are in bold. Random interventions are averaged over 10 seeds.}
\label{tab:main-results}
\centering
\setlength{\tabcolsep}{3.5pt}
\scalebox{0.865}{
\begin{tabular}{l|l|l|c|c|c|c|c}
\toprule \midrule
Method & Intervention & Evaluation
& VisA
& MVTec AD
& MVTec LOCO
& MVTec AD 2
& All \\
& &
& (12)
& (15)
& (5)
& (8)
& (40) \\
\midrule \midrule
\multicolumn{2}{l|}{\multirow{2}{*}{Original PatchCore}}
& Source
& 0.8953
& 0.9859
& 0.8430
& 0.6440
& 0.8724 \\
\multicolumn{2}{l|}{}
& Corrupted
& 0.7868
& 0.9311
& 0.7831
& 0.6177
& 0.8066 \\
\midrule
\multicolumn{2}{l|}{\multirow{2}{*}{PatchCore + extra normals}}
& Source
& 0.8955
& 0.9866
& 0.8482
& 0.6521
& 0.8751 \\
\multicolumn{2}{l|}{}
& Corrupted
& 0.7862
& 0.9316
& 0.7874
& 0.6236
& 0.8083 \\
\midrule
\multicolumn{2}{l|}{\multirow{2}{*}{DINOv3 linear probe}}
& Source
& 0.8979
& 0.8544
& 0.6199
& 0.5998
& 0.7872 \\
\multicolumn{2}{l|}{}
& Corrupted
& \textbf{0.8664}
& 0.8296
& 0.6076
& 0.6094
& 0.7688 \\
\midrule \midrule
\multirow{6}{*}{Random}
& \multirow{2}{*}{Suppression}
& Source
& 0.8903
& 0.9836
& 0.8390
& 0.6542
& 0.8716 \\
& & Corrupted
& 0.7722
& 0.9191
& 0.7814
& 0.6233
& 0.7987 \\
\cmidrule(l){2-8}
& \multirow{2}{*}{Amplification}
& Source
& 0.8593
& 0.9832
& 0.8397
& 0.6249
& 0.8564 \\
& & Corrupted
& 0.7632
& 0.9272
& 0.7856
& 0.6036
& 0.7956 \\
\cmidrule(l){2-8}
& \multirow{2}{*}{Joint}
& Source
& 0.8581
& 0.9761
& 0.8393
& 0.6304
& 0.8545 \\
& & Corrupted
& 0.7556
& 0.9128
& 0.7814
& 0.6100
& 0.7887 \\
\midrule \midrule
\multirow{6}{*}{Calibration-guided}
& \multirow{2}{*}{Suppression}
& Source
& 0.9006
& 0.9886
& 0.8490
& 0.6581
& 0.8787 \\
& & Corrupted
& 0.7928
& 0.9358
& 0.7938
& 0.6271
& 0.8134 \\
\cmidrule(l){2-8}
& \multirow{2}{*}{Amplification}
& Source
& 0.9052
& 0.9874
& 0.8554
& 0.6478
& 0.8783 \\
& & Corrupted
& 0.8057
& 0.9328
& \textbf{0.8000}
& 0.6162
& 0.8148 \\
\cmidrule(l){2-8}
& \multirow{2}{*}{Joint}
& Source
& 0.9061
& 0.9882
& \textbf{0.8570}
& 0.6579
& 0.8811 \\
& & Corrupted
& 0.8080
& 0.9371
& 0.7986
& 0.6269
& 0.8190 \\
\midrule \midrule
\multirow{6}{*}{AutoInterp-guided}
& \multirow{2}{*}{Suppression}
& Source
& 0.9080
& \textbf{0.9893}
& 0.8419
& 0.6612
& 0.8808 \\
& & Corrupted
& 0.7923
& 0.9352
& 0.7891
& \textbf{0.6276}
& 0.8126 \\
\cmidrule(l){2-8}
& \multirow{2}{*}{Amplification}
& Source
& 0.9198
& 0.9849
& 0.8400
& 0.6580
& 0.8819 \\
& & Corrupted
& 0.8134
& 0.9312
& 0.7867
& 0.6184
& 0.8152 \\
\cmidrule(l){2-8}
& \multirow{2}{*}{Joint}
& Source
& \textbf{0.9242}
& 0.9886
& 0.8419
& \textbf{0.6626}
& \textbf{0.8857} \\
& & Corrupted
& 0.8162
& \textbf{0.9389}
& 0.7891
& 0.6269
& \textbf{0.8210} \\
\midrule
\bottomrule
\end{tabular}
}
\end{table}
\paragraph{Evaluation Protocol.}
We report image-level AUROC and macro-average results, giving each category
equal weight. Results on VisA, MVTec AD, MVTec LOCO AD, and MVTec AD 2 are
macro-averaged across their 40 categories, while RobustAD is reported
separately across its three categories. We compare the original PatchCore with
two direct uses of the available calibration data and with three strategies for
assigning SAE features to interventions. First, we evaluate PatchCore with the normal calibration images included in memory-bank construction while retaining the original coreset ratio of 0.1. We also evaluate a class-weighted DINOv3 linear
probe on globally average-pooled patch representations from the same frozen
backbone. The probe is trained on known-normal training images and both normal
and anomalous calibration images. The purpose of these baselines is to test the effect of using the available
calibration data directly without any sparse feature decomposition or automated interpretability.
For feature interventions, we compare AutoInterp-guided, calibration-guided,
and random feature assignments, applying suppression, amplification, and joint
intervention under each strategy. For the calibration-guided assignment, we
compute the calibration AUROC of each SAE feature, treating features below
$0.5$ as distractors and those above $0.5$ as anomaly features, with ties left
unchanged. This provides an alternative protocol testing whether the AutoInterp procedure adds value beyond assigning feature labels directly from
calibration supervision.
For the random assignment, we sample feature sets matching the numbers of
AutoInterp-labeled distractor and anomaly features for each category. The two
sets are disjoint for the joint intervention. Random amplification and joint
intervention use the corresponding category-level amplification factor selected
by the AutoInterp procedure. This control tests whether performance depends on
which SAE features are selected for suppression and amplification, while keeping
the number of selected features and the corresponding amplification multiplier
fixed.

To evaluate robustness to synthetic image corruptions, we apply brightness,
contrast, defocus blur, and Gaussian noise from
\href{https://github.com/bethgelab/imagecorruptions}{\texttt{imagecorruptions}}
at severities $1$--$5$
\citep{hendrycks2018benchmarking,michaelis2019dragon}. We additionally evaluate
robustness to real acquisition shifts on RobustAD. In both evaluations, feature
assignments and amplification factors are determined using only source
calibration data and remain fixed across the corrupted or shifted evaluation
data.

\section{Results and Discussion}
\label{sec:results}

\paragraph{RQ1: Performance on source data.}
Table~\ref{tab:main-results} reports performance across the 40
categories of VisA, MVTec AD, MVTec LOCO AD, and MVTec AD 2.
On source data, AutoInterp-guided interventions improve the macro-average AUROC of the original
PatchCore from 0.8724 to 0.8808 with distractor suppression, 0.8819 with anomaly
amplification, and 0.8857 with their joint intervention. The joint intervention
improves performance on three of the four dataset families, with MVTec LOCO AD
being the exception. Improvements are also observed when the original detector
is already close to saturation. For example, on MVTec AD, PatchCore reaches
0.9859 AUROC, while AutoInterp-guided suppression reaches 0.9893.

The controls suggest that these gains are not explained by simply using the calibration data. Adding normal
calibration images to PatchCore memory bank improves overall AUROC by only
0.27 points. A linear probe trained on all normal training images plus the calibration set
performs substantially worse, reaching 0.7872 AUROC. This may partly reflect the limited amount of supervised data available,
with a median of only 10 anomalous calibration examples per category.
Calibration-guided interventions make more effective use of the same
calibration supervision, with joint intervention reaching 0.8811 and
outperforming AutoInterp on MVTec LOCO AD, but remaining below AutoInterp
overall. Among the intervention procedures, as expected, random feature
assignment gives the weakest results overall, but notably does not collapse the
detector. We initially hypothesized that this tolerance to arbitrary edits might
arise from the substantial signal retained in the reconstruction error alone.
However, this tolerance persists when the reconstruction error is omitted
(see Appendix~\ref{app:sae}).

\begin{table}[t]
\caption{Image-level AUROC on RobustAD source and acquisition shifts,
macro-averaged across categories. Best results are in bold. Random
interventions are averaged over 10 seeds.}
\label{tab:robustad-results}
\centering
\setlength{\tabcolsep}{3.5pt}
\scalebox{0.865}{
\begin{tabular}{l|l|l|c|c|c|c}
\toprule \midrule
Method & Intervention & Evaluation
& PCB
& MetalParts
& PiledBags
& All (3) \\
\midrule \midrule
\multicolumn{2}{l|}{\multirow{2}{*}{Original PatchCore}}
& Source
& 0.8926
& 0.7423
& 0.9887
& 0.8745 \\
\multicolumn{2}{l|}{}
& Shifted
& 0.3943
& 0.6415
& 0.7849
& 0.6069 \\
\midrule
\multicolumn{2}{l|}{\multirow{2}{*}{PatchCore + extra normals}}
& Source
& 0.9202
& 0.7386
& 0.9889
& 0.8826 \\
\multicolumn{2}{l|}{}
& Shifted
& 0.4154
& 0.6413
& 0.7828
& 0.6131 \\
\midrule
\multicolumn{2}{l|}{\multirow{2}{*}{DINOv3 linear probe}}
& Source
& \textbf{0.9998}
& 0.7495
& 0.8057
& 0.8517 \\
\multicolumn{2}{l|}{}
& Shifted
& 0.5165
& \textbf{0.7106}
& 0.7177
& 0.6482 \\
\midrule \midrule
\multirow{6}{*}{Random}
& \multirow{2}{*}{Suppression}
& Source
& 0.8587
& 0.7572
& 0.9652
& 0.8604 \\
& & Shifted
& 0.4271
& 0.6363
& 0.7424
& 0.6019 \\
\cmidrule(l){2-7}
& \multirow{2}{*}{Amplification}
& Source
& 0.9509
& 0.7193
& 0.9774
& 0.8825 \\
& & Shifted
& 0.4368
& 0.6402
& 0.7994
& 0.6254 \\
\cmidrule(l){2-7}
& \multirow{2}{*}{Joint}
& Source
& 0.9495
& 0.7573
& 0.9651
& 0.8906 \\
& & Shifted
& 0.4401
& 0.6363
& 0.7423
& 0.6062 \\
\midrule \midrule
\multirow{6}{*}{Calibration-guided}
& \multirow{2}{*}{Suppression}
& Source
& 0.8946
& 0.7479
& \textbf{0.9993}
& 0.8806 \\
& & Shifted
& 0.3956
& 0.6424
& 0.7769
& 0.6049 \\
\cmidrule(l){2-7}
& \multirow{2}{*}{Amplification}
& Source
& 0.9722
& 0.6587
& 0.9990
& 0.8766 \\
& & Shifted
& 0.2928
& 0.6269
& 0.8247
& 0.5815 \\
\cmidrule(l){2-7}
& \multirow{2}{*}{Joint}
& Source
& 0.9722
& 0.7015
& \textbf{0.9993}
& 0.8910 \\
& & Shifted
& 0.2891
& 0.6255
& 0.7769
& 0.5638 \\
\midrule \midrule
\multirow{6}{*}{AutoInterp-guided}
& \multirow{2}{*}{Suppression}
& Source
& 0.8822
& \textbf{0.7747}
& 0.9958
& 0.8842 \\
& & Shifted
& 0.4899
& 0.6574
& 0.8016
& 0.6496 \\
\cmidrule(l){2-7}
& \multirow{2}{*}{Amplification}
& Source
& 0.9456
& 0.7374
& 0.9921
& 0.8917 \\
& & Shifted
& 0.5149
& 0.6568
& \textbf{0.8452}
& \textbf{0.6723} \\
\cmidrule(l){2-7}
& \multirow{2}{*}{Joint}
& Source
& 0.9464
& \textbf{0.7747}
& 0.9958
& \textbf{0.9056} \\
& & Shifted
& \textbf{0.5208}
& 0.6574
& 0.8016
& 0.6599 \\
\midrule
\bottomrule
\end{tabular}
}
\end{table}

\paragraph{RQ2: Robustness to corruptions and acquisition shifts.}
We next test whether the improvements obtained on source data persist when the
input distribution changes. Under synthetic corruptions, the joint
AutoInterp-guided intervention improves macro-average AUROC from 0.8066 to
0.8210, a gain of 1.44 points, slightly larger than the 1.33-point gain observed
on source data. The improvement persists at every corruption severity, with the
largest gap at severity 5. At this severity, joint intervention improves AUROC
by 1.83 points on average and outperforms the original PatchCore in 30 of the
40 categories. On RobustAD, the improvement is particularly large under real acquisition
shifts. Joint intervention improves source AUROC from 0.8745 to 0.9056,
a gain of 3.11 points, while under shifted acquisition conditions it improves
AUROC from 0.6069 to 0.6599, a gain of 5.30 points. The individual suppression
and amplification interventions show the same qualitative pattern, with larger
gains under shifted acquisition conditions than on source data. RobustAD also
provides substantially more calibration data than the other benchmarks, which
could contribute to better SAE estimation. Its calibration sets contain a
median of 150 images per category, compared with a median of 20 images per
category across the 40-category benchmark. However, within those 40 categories,
whose calibration sets range from 4 to 54 images, calibration-set size shows
only a modest positive association with joint-intervention gain on source data
(Spearman's $\rho=0.371$), which becomes substantially weaker under corruptions
($\rho=0.191$). 

Regarding the calibration-guided control, the joint intervention is competitive
under synthetic corruptions, reaching 0.8190 AUROC compared with 0.8210 for
AutoInterp. Under the real acquisition shifts of RobustAD, however, it reaches
0.5638, compared with 0.6069 for the original PatchCore and 0.6599 for
AutoInterp. Since
AutoInterp and calibration guidance use the same calibration data and SAE,
their different behavior suggests that semantic feature interpretation provides
information beyond calibration AUROC, especially under real acquisition shifts.

The linear probe shows a heterogeneous pattern across
both robustness evaluations. Under synthetic corruptions, it substantially
outperforms the original PatchCore on VisA but falls below it on MVTec AD,
MVTec LOCO AD, and MVTec AD 2. Under the real acquisition shifts of RobustAD, it
outperforms PatchCore on PCB and MetalParts but falls below it on PiledBags.
This heterogeneous behavior suggests that direct prediction from the
calibration data may be highly sensitive to how well the available examples
represent the evaluation distribution.

\begin{figure}[!t]
    \centering
    \includegraphics[width=1\linewidth]{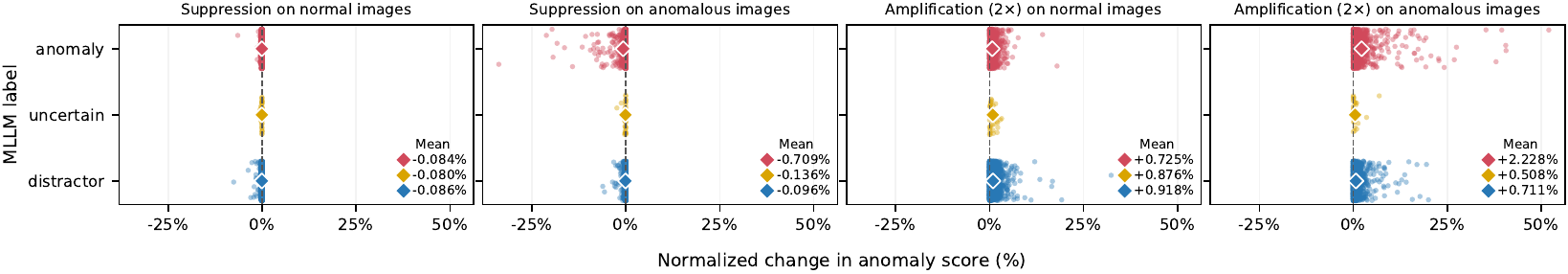}
\caption{
Effects of suppressing and amplifying SAE features across 43 categories.
Points show normalized mean score changes per feature, and diamonds show
category-balanced means by label.
}
\label{fig:feature-effects}
\end{figure}

\paragraph{RQ3: Analysis of individual feature interventions by MLLM label.}
We next test whether the MLLM labels predict the class-specific effects of
individual SAE feature interventions on PatchCore scores. For each non-dead
feature, we independently suppress or amplify it across all patches of every
test image, rerun PatchCore, and measure the resulting change in image anomaly
score separately for normal and anomalous images. Score changes are normalized
by the category's mean original test score, computed over all test images. We use a fixed $2\times$ amplification factor, close to
the mean automatically selected amplification multiplier of $1.884\times$
across the 43 categories. Of the 64 SAE features per category, an average of
63.86 are non-dead, with 22.44 labeled as anomaly, 40.58 as distractor, and
0.84 as uncertain. In Figure~\ref{fig:feature-effects}, each point represents
the average score change produced by one feature across all images of the
corresponding class. Consistent with what we would expect from anomaly and distractor features,
suppressing anomaly features reduces scores more on anomalous than on
normal images, while amplifying them increases scores more on anomalous images.
In contrast, distractor features have similar effects on the two
classes.

\newcommand{\imgcell}[1]{%
  \begin{minipage}[t]{0.235\linewidth}\centering
    \includegraphics[width=\linewidth]{#1}
  \end{minipage}}

\begin{figure*}[t]
\centering

\begin{minipage}[t]{0.487\textwidth}
\centering
{\small (a) Feature labeled as anomaly}\\[-0.35mm]
{\scriptsize RobustAD PiledBags, SAE feature 38}\\[-0.25mm]
{\footnotesize\itshape ``upside-down (flipped) packet label''}\\[1.15mm]

\begin{minipage}[t]{0.235\linewidth}\centering
{\scriptsize Test image}
\end{minipage}\hfill
\begin{minipage}[t]{0.745\linewidth}\centering
{\scriptsize Top-activating calibration examples}
\end{minipage}\\[0.55mm]

\makebox[\linewidth][c]{%
\imgcell{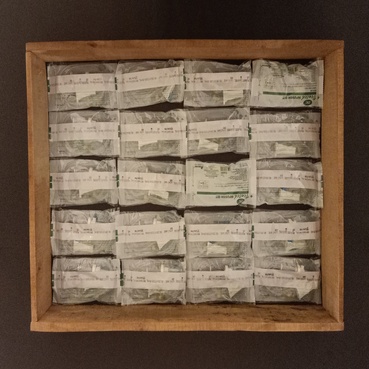}\hfill
\imgcell{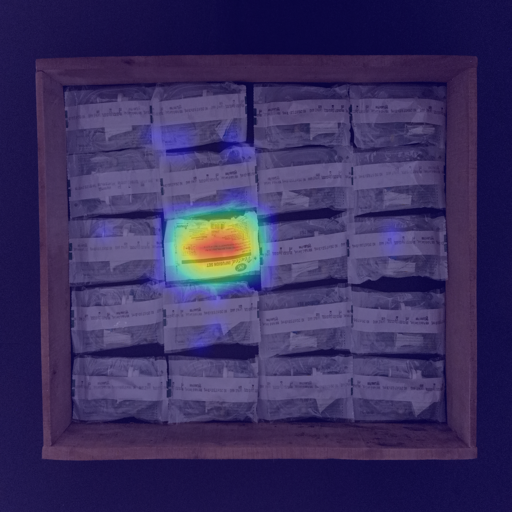}\hfill
\imgcell{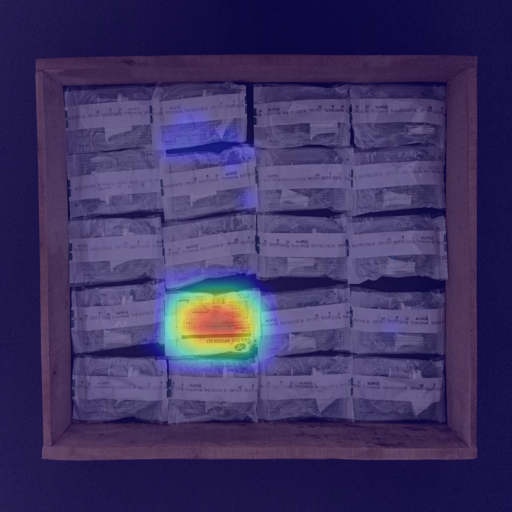}\hfill
\imgcell{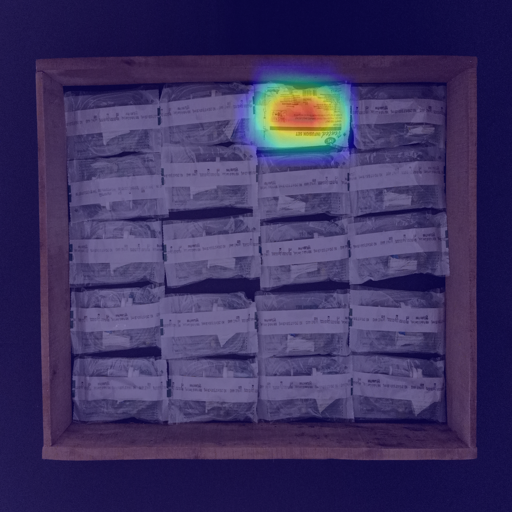}}\\[1.35mm]

\begin{minipage}[t]{0.235\linewidth}\centering
{\scriptsize\mbox{SAE feature map}}
\end{minipage}\hfill
\begin{minipage}[t]{0.745\linewidth}\centering
{\scriptsize PatchCore anomaly maps}
\end{minipage}\\[0.55mm]

\makebox[\linewidth][c]{%
\imgcell{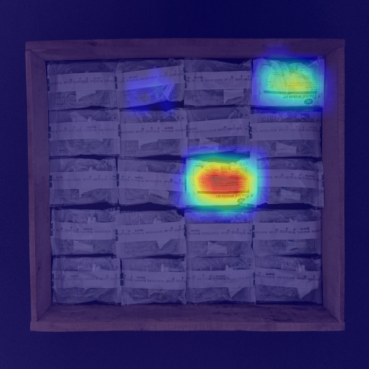}\hfill
\imgcell{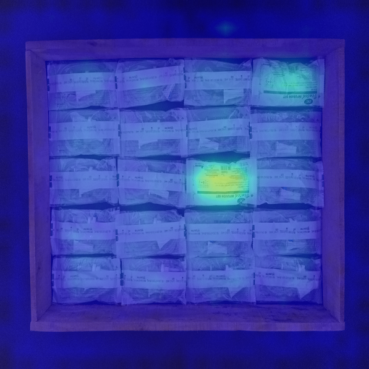}\hfill
\imgcell{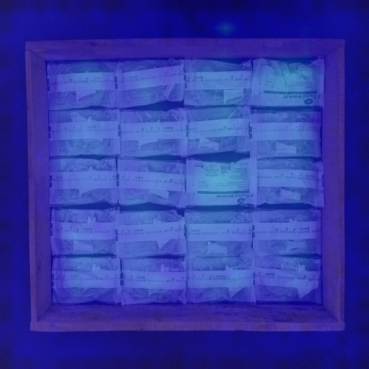}\hfill
\imgcell{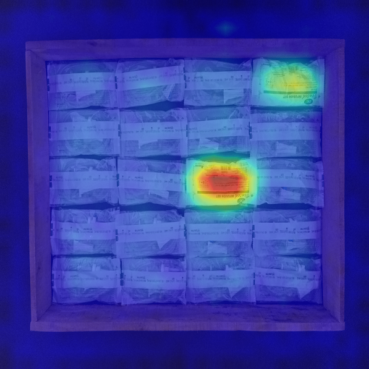}}\\[-0.1mm]

\begin{minipage}[t]{0.235\linewidth}\centering~\end{minipage}\hfill
\begin{minipage}[t]{0.235\linewidth}\centering
{\scriptsize Original}
\end{minipage}\hfill
\begin{minipage}[t]{0.235\linewidth}\centering
{\scriptsize Suppressed}
\end{minipage}\hfill
\begin{minipage}[t]{0.235\linewidth}\centering
{\scriptsize Amplified}
\end{minipage}
\end{minipage}%
\hfill%
\begin{minipage}[t]{0.487\textwidth}
\centering
{\small (b) Feature labeled as distractor}\\[-0.35mm]
{\scriptsize MVTec AD 2 Can, SAE feature 5}\\[-0.25mm]
{\footnotesize\itshape ``localized glossy glare or specular highlight''}\\[1.15mm]

\begin{minipage}[t]{0.235\linewidth}\centering
{\scriptsize Test image}
\end{minipage}\hfill
\begin{minipage}[t]{0.745\linewidth}\centering
{\scriptsize Top-activating calibration examples}
\end{minipage}\\[0.55mm]

\makebox[\linewidth][c]{%
\imgcell{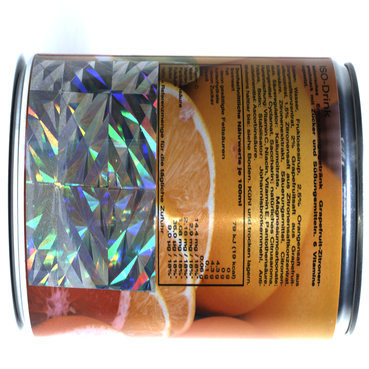}\hfill
\imgcell{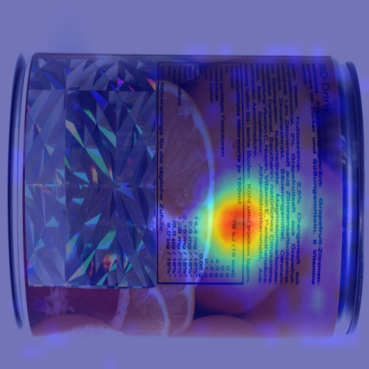}\hfill
\imgcell{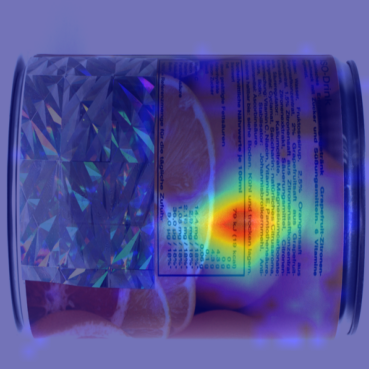}\hfill
\imgcell{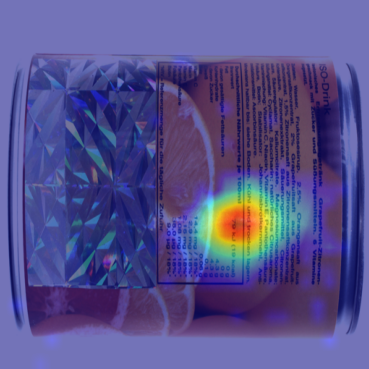}}\\[1.35mm]

\begin{minipage}[t]{0.235\linewidth}\centering
{\scriptsize\mbox{SAE feature map}}
\end{minipage}\hfill
\begin{minipage}[t]{0.745\linewidth}\centering
{\scriptsize PatchCore anomaly maps}
\end{minipage}\\[0.55mm]

\makebox[\linewidth][c]{%
\imgcell{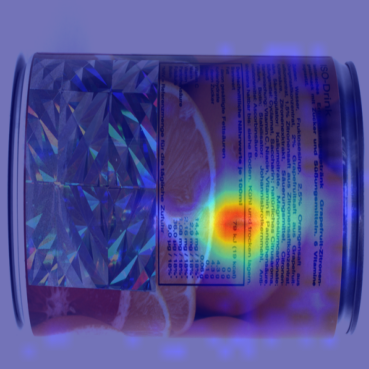}\hfill
\imgcell{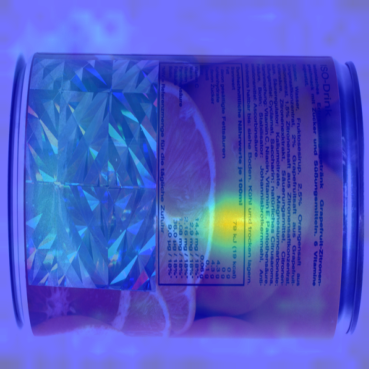}\hfill
\imgcell{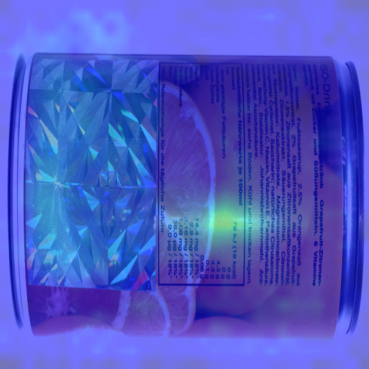}\hfill
\imgcell{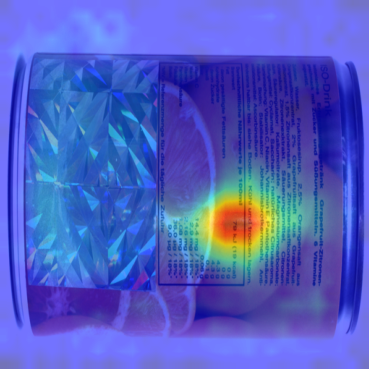}}\\[-0.1mm]

\begin{minipage}[t]{0.235\linewidth}\centering~\end{minipage}\hfill
\begin{minipage}[t]{0.235\linewidth}\centering
{\scriptsize Original}
\end{minipage}\hfill
\begin{minipage}[t]{0.235\linewidth}\centering
{\scriptsize Suppressed}
\end{minipage}\hfill
\begin{minipage}[t]{0.235\linewidth}\centering
{\scriptsize Amplified}
\end{minipage}
\end{minipage}

\caption{
Examples of anomaly and distractor SAE features.
Italic text gives the MLLM-generated feature description. The top row shows a
test image and three top-activating calibration examples shown to the MLLM.
The bottom row shows, for the test image, the SAE feature map and PatchCore
anomaly maps before intervention, after suppression, and after amplification.
The three PatchCore anomaly maps for each example share a min-max color scale.
}
\label{fig:qualitative-features}
\end{figure*}
To assess whether these
class-specific differences are consistent across the 43 categories, we perform
four statistical tests, one for each combination of feature label and
intervention. For each test, we compare category-level average score
changes on anomalous images with corresponding averages on normal images.
For anomaly features, we use one-sided paired $t$-tests in the predicted
direction. For distractor features, we use two-sided paired $t$-tests
because no directional class difference is predicted. We apply Holm correction
across the four tests. For anomaly features, the class-specific
difference is significant under both suppression
($p_{\mathrm{Holm}}=2.07\times10^{-6}$) and amplification
($p_{\mathrm{Holm}}=1.73\times10^{-6}$), supporting the pattern observed in
Figure~\ref{fig:feature-effects}. For distractor features, the
difference is not significant under either suppression
($p_{\mathrm{Holm}}=0.346$) or amplification
($p_{\mathrm{Holm}}=0.0755$), with distractor amplification showing a larger increase on normal images. Uncertain features are excluded from these tests because they are few
and occur in only 15 of 43 categories.

Another pattern visible in Figure~\ref{fig:feature-effects} is that the
separation between features labeled as anomaly and features labeled as
distractor is concentrated toward the tails of the distributions. At the 10th
percentile for suppression and the 90th percentile for amplification, the
difference between anomaly and distractor score changes is 1.71 and 2.03 times
the corresponding average difference across all features, respectively.
Figure~\ref{fig:qualitative-features} provides qualitative examples of one
feature labeled as anomaly and one labeled as distractor, together with their
effects on PatchCore anomaly maps. In the shown examples, suppressing the
PiledBags feature described as \textit{``upside-down (flipped) packet label''}
reduces the score of the anomalous image by approximately $42.3\%$, while
suppressing the Can feature described as \textit{``localized glossy glare or
specular highlight''} reduces the score of the normal image by $29.32\%$.
These examples show much larger score changes than the average effects in
Figure~\ref{fig:feature-effects}, as the latter averages the effect over all
images of the corresponding class, including many where the feature has little
or no effect. More qualitative examples are provided in
Appendix~\ref{app:qualitative}.

\section{Conclusion and Limitations}
We presented an approach for augmenting visual anomaly detectors by decomposing
their anomaly signal into sparse features, automatically interpreting these
features with an MLLM, and using the resulting labels to guide interventions on
the detector. Across a total of 43 categories from five benchmarks, these
interventions improve PatchCore performance on source data and robustness under
distribution shift. The gains persist under synthetic corruptions and are
particularly large under the real acquisition shifts of RobustAD. At the
feature level, we further find that the MLLM labels are aligned in aggregate
with how the corresponding features affect normal and anomalous images.
Together, these results show that automatically interpreted sparse features can
provide an actionable interface to the anomaly signal of a frozen detector.

Our study also has important limitations. First, the method requires a calibration
set containing anomalous examples, and the amplification factor is selected
using calibration AUROC. The intervention procedure is therefore not purely
normal-only, even when applied to an unsupervised anomaly detector. Second,
although the interventions improve performance on average, they do not
guarantee an improvement for every category or model configuration and can
degrade the original detector in some cases. Finally, the generated feature descriptions and labels should be viewed as
hypotheses about feature semantics, since the MLLM may misinterpret the provided
examples and individual SAE features may themselves remain polysemantic. Our
feature-level analysis shows that their behavior is consistent with the assigned
labels in aggregate, but there is no ground-truth annotation for
individual SAE features and therefore no guarantee that each interpretation is
faithful. 

\subsection*{AI use statement}
In this work, we used generative AI tools to propose or refine hypotheses,
provide feedback on research methodology and experiments, implement methods,
and interpret results. We have not used generative AI tools to generate
synthetic datasets, help develop theoretical models or conceptual frameworks,
or clean and reformat datasets. Formulating mathematical claims, providing
critical ingredients for proving mathematical claims, assisting in the writing
of proofs, assisting with translation, and supporting qualitative and thematic
data analysis are not applicable to this work.
Additionally, we used generative AI tools to create or edit software code,
draft parts of the research paper, brainstorm, and edit the research paper to
improve readability.
We have reviewed all AI-assisted work. LLM-generated code was reviewed and
tested for correctness, and LLM-generated manuscript content was heavily edited
by the authors.
We take responsibility for the final content of this work, including text,
claims or artifacts produced with the aid of generative AI.

\bibliography{iclr2027_conference}

@INPROCEEDINGS{patchcore,
  author={Roth, Karsten and Pemula, Latha and Zepeda, Joaquin and Schölkopf, Bernhard and Brox, Thomas and Gehler, Peter},
  booktitle={2022 IEEE/CVF Conference on Computer Vision and Pattern Recognition (CVPR)}, 
  title={Towards Total Recall in Industrial Anomaly Detection}, 
  year={2022},
  volume={},
  number={},
  pages={14298-14308},
  doi={10.1109/CVPR52688.2022.01392}}

@inproceedings{fre,
author    = {Ibrahima Ndiour and Nilesh A Ahuja and Ergin U Genc and Omesh Tickoo},
title     = {FRE: A Fast Method For Anomaly Detection And Segmentation},
booktitle = {34th British Machine Vision Conference 2023, {BMVC} 2023, Aberdeen, UK, November 20-24, 2023},
publisher = {BMVA},
year      = {2023},
url       = {https://papers.bmvc2023.org/0614.pdf}
}

@inproceedings{
gao2025scaling,
title={Scaling and evaluating sparse autoencoders},
author={Leo Gao and Tom Dupre la Tour and Henk Tillman and Gabriel Goh and Rajan Troll and Alec Radford and Ilya Sutskever and Jan Leike and Jeffrey Wu},
booktitle={The Thirteenth International Conference on Learning Representations},
year={2025},
url={https://openreview.net/forum?id=tcsZt9ZNKD}
}

@misc{bricken2023monosemanticity,
  title = {Towards Monosemanticity: Decomposing Language Models With Dictionary Learning},
  author = {Bricken, Trenton and Templeton, Adly and Chen, Ben and Lindsey, Jack and Jermyn, Adam and Carter, Shan and Henighan, Tom and Pearce, Adam and Olah, Chris},
  year = {2023},
  howpublished = {\url{https://transformer-circuits.pub/2023/monosemantic-features}},
  note = {Transformer Circuits Thread, Anthropic}
}

@inproceedings{
hendrycks2018benchmarking,
title={Benchmarking Neural Network Robustness to Common Corruptions and Perturbations},
author={Dan Hendrycks and Thomas Dietterich},
booktitle={International Conference on Learning Representations},
year={2019},
url={https://openreview.net/forum?id=HJz6tiCqYm},
}

@article{michaelis2019dragon,
  title={Benchmarking Robustness in Object Detection: 
    Autonomous Driving when Winter is Coming},
  author={Michaelis, Claudio and Mitzkus, Benjamin and 
    Geirhos, Robert and Rusak, Evgenia and 
    Bringmann, Oliver and Ecker, Alexander S. and 
    Bethge, Matthias and Brendel, Wieland},
  journal={arXiv preprint arXiv:1907.07484},
  year={2019}
}

@misc{bills2023language,
 title={Language models can explain neurons in language models},
 author={
    Bills, Steven and Cammarata, Nick and Mossing, Dan and Tillman, Henk and Gao, Leo and Goh, Gabriel and Sutskever, Ilya and Leike, Jan and Wu, Jeff and Saunders, William
 },
 year={2023},
 howpublished = {\url{https://openaipublic.blob.core.windows.net/neuron-explainer/paper/index.html}}
}

@inproceedings{shaham2024multimodal,
	  title={A multimodal automated interpretability agent},
	  author={Rott Shaham, Tamar and Schwettmann, Sarah and Wang, Franklin and Rajaram, Achyuta and Hernandez, Evan and Andreas, Jacob and Torralba, Antonio},
	  booktitle={Forty-first International Conference on Machine Learning},
	  year={2024}
	}

@article{templeton2024scaling,
   title={Scaling Monosemanticity: Extracting Interpretable Features from Claude 3 Sonnet},
   author={Templeton, Adly and Conerly, Tom and Marcus, Jonathan and Lindsey, Jack and Bricken, Trenton and Chen, Brian and Pearce, Adam and Citro, Craig and Ameisen, Emmanuel and Jones, Andy and Cunningham, Hoagy and Turner, Nicholas L and McDougall, Callum and MacDiarmid, Monte and Freeman, C. Daniel and Sumers, Theodore R. and Rees, Edward and Batson, Joshua and Jermyn, Adam and Carter, Shan and Olah, Chris and Henighan, Tom},
   year={2024},
   journal={Transformer Circuits Thread},
   url={https://transformer-circuits.pub/2024/scaling-monosemanticity/index.html}
}

@inproceedings{
desantis2026learning,
title={Learning Concept Bottleneck Models from Mechanistic Explanations},
author={Antonio {De Santis} and Schrasing Tong and Marco Brambilla and Lalana Kagal},
booktitle={The Fourteenth International Conference on Learning Representations},
year={2026},
url={https://openreview.net/forum?id=gdEWoxhb70}
}

@article{olah2020zoom,
  author = {Olah, Chris and Cammarata, Nick and Schubert, Ludwig and Goh, Gabriel and Petrov, Michael and Carter, Shan},
  title = {Zoom In: An Introduction to Circuits},
  journal = {Distill},
  year = {2020},
  note = {https://distill.pub/2020/circuits/zoom-in},
  doi = {10.23915/distill.00024.001}
}

@inproceedings{netdissect2017,
  title={Network Dissection: Quantifying Interpretability of Deep Visual Representations},
  author={Bau, David and Zhou, Bolei and Khosla, Aditya and Oliva, Aude and Torralba, Antonio},
  booktitle={Computer Vision and Pattern Recognition},
  year={2017}
}

@InProceedings{kim2018interpretability,
  title = 	 {Interpretability Beyond Feature Attribution: Quantitative Testing with Concept Activation Vectors ({TCAV})},
  author =       {Kim, Been and Wattenberg, Martin and Gilmer, Justin and Cai, Carrie and Wexler, James and Viegas, Fernanda and sayres, Rory},
  booktitle = 	 {Proceedings of the 35th International Conference on Machine Learning},
  pages = 	 {2668--2677},
  year = 	 {2018},
  editor = 	 {Dy, Jennifer and Krause, Andreas},
  volume = 	 {80},
  series = 	 {Proceedings of Machine Learning Research},
  month = 	 {10--15 Jul},
  publisher =    {PMLR},
  url = 	 {https://proceedings.mlr.press/v80/kim18d.html}
}

@article{
desantis2026visualtcav,
title={Visual-{TCAV}: Concept-based Attribution and Saliency Maps for Post-hoc Explainability in Image Classification},
author={Antonio {De Santis} and Riccardo Campi and Matteo Bianchi and Marco Brambilla},
journal={Transactions on Machine Learning Research},
issn={2835-8856},
year={2026},
url={https://openreview.net/forum?id=SLh00W5rhu},
note={}
}

@inproceedings{ace,
 author = {Ghorbani, Amirata and Wexler, James and Zou, James Y and Kim, Been},
 booktitle = {Advances in Neural Information Processing Systems},
 editor = {H. Wallach and H. Larochelle and A. Beygelzimer and F. d\textquotesingle Alch\'{e}-Buc and E. Fox and R. Garnett},
 pages = {},
 publisher = {Curran Associates, Inc.},
 title = {Towards Automatic Concept-based Explanations},
 volume = {32},
 year = {2019}
}

@article{ice, title={Invertible Concept-based Explanations for CNN Models with Non-negative Concept Activation Vectors}, volume={35}, url={https://ojs.aaai.org/index.php/AAAI/article/view/17389}, DOI={10.1609/aaai.v35i13.17389}, abstractNote={Convolutional neural network (CNN) models for computer vision are powerful but lack explainability in their most basic form. This deficiency remains a key challenge when applying CNNs in important domains. Recent work on explanations through feature importance of approximate linear models has moved from input-level features (pixels or segments) to features from mid-layer feature maps in the form of concept activation vectors (CAVs). CAVs contain concept-level information and could be learned via clustering. In this work, we rethink the ACE algorithm of Ghorbani et~al., proposing an alternative invertible concept-based explanation (ICE) framework to overcome its shortcomings. Based on the requirements of fidelity (approximate models to target models) and interpretability (being meaningful to people), we design measurements and evaluate a range of matrix factorization methods with our framework. We find that non-negative concept activation vectors (NCAVs) from non-negative matrix factorization provide superior performance in interpretability and fidelity based on computational and human subject experiments. Our framework provides both local and global concept-level explanations for pre-trained CNN models.}, number={13}, journal={Proceedings of the AAAI Conference on Artificial Intelligence}, author={Zhang, Ruihan and Madumal, Prashan and Miller, Tim and Ehinger, Krista A. and Rubinstein, Benjamin I. P.}, year={2021}, month={May}, pages={11682-11690} }

@misc{elhage2022toymodelssuperposition,
      title={Toy Models of Superposition}, 
      author={Nelson Elhage and Tristan Hume and Catherine Olsson and Nicholas Schiefer and Tom Henighan and Shauna Kravec and Zac Hatfield-Dodds and Robert Lasenby and Dawn Drain and Carol Chen and Roger Grosse and Sam McCandlish and Jared Kaplan and Dario Amodei and Martin Wattenberg and Christopher Olah},
      year={2022},
      eprint={2209.10652},
      archivePrefix={arXiv},
      primaryClass={cs.LG},
      url={https://arxiv.org/abs/2209.10652}, 
}

@inproceedings{rimsky-etal-2024-steering,
    title = "Steering Llama 2 via Contrastive Activation Addition",
    author = "Rimsky, Nina  and
      Gabrieli, Nick  and
      Schulz, Julian  and
      Tong, Meg  and
      Hubinger, Evan  and
      Turner, Alexander",
    editor = "Ku, Lun-Wei  and
      Martins, Andre  and
      Srikumar, Vivek",
    booktitle = "Proceedings of the 62nd Annual Meeting of the Association for Computational Linguistics (Volume 1: Long Papers)",
    month = aug,
    year = "2024",
    address = "Bangkok, Thailand",
    publisher = "Association for Computational Linguistics",
    url = "https://aclanthology.org/2024.acl-long.828/",
    doi = "10.18653/v1/2024.acl-long.828",
    pages = "15504--15522"
}

@InProceedings{craft,
    author    = {Fel, Thomas and Picard, Agustin and B\'ethune, Louis and Boissin, Thibaut and Vigouroux, David and Colin, Julien and Cad\`ene, R\'emi and Serre, Thomas},
    title     = {CRAFT: Concept Recursive Activation FacTorization for Explainability},
    booktitle = {Proceedings of the IEEE/CVF Conference on Computer Vision and Pattern Recognition (CVPR)},
    month     = {June},
    year      = {2023},
    pages     = {2711-2721}
}

@InProceedings{abc,
    author    = {Bianchi, Matteo and Campi, Riccardo and De Santis, Antonio and Merengo, Sara and Brambilla, Marco},
    title     = {Activation-Based Concept Extraction for Explainability in Image Classification},
    booktitle = {Proceedings of the IEEE/CVF Conference on Computer Vision and Pattern Recognition (CVPR) Workshops},
    month     = {June},
    year      = {2026},
    pages     = {3996-4005}
}

@inproceedings{
    marks2025sparse,
    title={Sparse Feature Circuits: Discovering and Editing Interpretable Causal Graphs in Language Models},
    author={Samuel Marks and Can Rager and Eric J Michaud and Yonatan Belinkov and David Bau and Aaron Mueller},
    booktitle={The Thirteenth International Conference on Learning Representations},
    year={2025},
    url={https://openreview.net/forum?id=I4e82CIDxv}
}

@inproceedings{padim,
author = {Defard, Thomas and Setkov, Aleksandr and Loesch, Angelique and Audigier, Romaric},
title = {PaDiM: A Patch Distribution Modeling Framework for Anomaly Detection and Localization},
year = {2021},
isbn = {978-3-030-68798-4},
publisher = {Springer-Verlag},
address = {Berlin, Heidelberg},
url = {https://doi.org/10.1007/978-3-030-68799-1_35},
doi = {10.1007/978-3-030-68799-1_35},
booktitle = {Pattern Recognition. ICPR International Workshops and Challenges: Virtual Event, January 10–15, 2021, Proceedings, Part IV},
pages = {475–489},
numpages = {15}
}

@InProceedings{cflow,
    author    = {Gudovskiy, Denis and Ishizaka, Shun and Kozuka, Kazuki},
    title     = {CFLOW-AD: Real-Time Unsupervised Anomaly Detection With Localization via Conditional Normalizing Flows},
    booktitle = {Proceedings of the IEEE/CVF Winter Conference on Applications of Computer Vision (WACV)},
    month     = {January},
    year      = {2022},
    pages     = {98-107}
}

@InProceedings{reverse_distillation,
    author    = {Deng, Hanqiu and Li, Xingyu},
    title     = {Anomaly Detection via Reverse Distillation From One-Class Embedding},
    booktitle = {Proceedings of the IEEE/CVF Conference on Computer Vision and Pattern Recognition (CVPR)},
    month     = {June},
    year      = {2022},
    pages     = {9737-9746}
}

@InProceedings{efficientad,
    author    = {Batzner, Kilian and Heckler, Lars and K\"onig, Rebecca},
    title     = {EfficientAD: Accurate Visual Anomaly Detection at Millisecond-Level Latencies},
    booktitle = {Proceedings of the IEEE/CVF Winter Conference on Applications of Computer Vision (WACV)},
    month     = {January},
    year      = {2024},
    pages     = {128-138}
}

@misc{wang2025unveiling,
      title={Unveiling the Unseen: A Comprehensive Survey on Explainable Anomaly Detection in Images and Videos}, 
      author={Yizhou Wang and Dongliang Guo and Sheng Li and Octavia Camps and Yun Fu},
      year={2025},
      eprint={2302.06670},
      archivePrefix={arXiv},
      primaryClass={cs.LG},
      url={https://arxiv.org/abs/2302.06670}, 
}

@inproceedings{
liznerski2021explainable,
title={Explainable Deep One-Class Classification},
author={Philipp Liznerski and Lukas Ruff and Robert A. Vandermeulen and Billy Joe Franks and Marius Kloft and Klaus Robert Muller},
booktitle={International Conference on Learning Representations},
year={2021},
url={https://openreview.net/forum?id=A5VV3UyIQz}
}

@ARTICLE{jiang2023interpretability,
  author={Jiang, Rui and Xue, Yijia and Zou, Dongmian},
  journal={IEEE Access}, 
  title={Interpretability-Aware Industrial Anomaly Detection Using Autoencoders}, 
  year={2023},
  volume={11},
  number={},
  pages={60490-60500},
  doi={10.1109/ACCESS.2023.3286548}}

@Article{mvtec-c,
title = {Robustness benchmark for unsupervised anomaly detection models},
journal = {JUSTC},
volume = {54},
number = {1},
pages = {0103-1-0103-11},
year = {2024},
issn = {2097-7387},
doi = {10.52396/JUSTC-2022-0165},	
url = {https://justc.ustc.edu.cn/en/article/doi/10.52396/JUSTC-2022-0165},
author = {Pei Wang and Wei Zhai and Yang Cao}
}

@InProceedings{robustad,
    author    = {Pemula, Latha and Zhang, Dongqing and Dabeer, Onkar},
    title     = {Robust AD: A Real World Benchmark Dataset For Robustness in Industrial Anomaly Detection},
    booktitle = {Proceedings of the IEEE/CVF Conference on Computer Vision and Pattern Recognition (CVPR) Workshops},
    month     = {June},
    year      = {2025},
    pages     = {4086-4096}
}

@inproceedings{zou2022spot,
  title={SPot-the-Difference Self-supervised Pre-training for Anomaly Detection and Segmentation},
  author={Zou, Yang and Jeong, Jongheon and Pemula, Latha and Zhang, Dongqing and Dabeer, Onkar},
  booktitle={European Conference on Computer Vision},
  pages={392--408},
  year={2022},
  organization={Springer}
}

@InProceedings{bergmann2019mvtec,
author = {Bergmann, Paul and Fauser, Michael and Sattlegger, David and Steger, Carsten},
title = {MVTec AD -- A Comprehensive Real-World Dataset for Unsupervised Anomaly Detection},
booktitle = {Proceedings of the IEEE/CVF Conference on Computer Vision and Pattern Recognition (CVPR)},
month = {June},
year = {2019}
}

@article{bergmann2022beyond,
author = {Bergmann, Paul and Batzner, Kilian and Fauser, Michael and Sattlegger, David and Steger, Carsten},
title = {Beyond Dents and Scratches: Logical Constraints in Unsupervised Anomaly Detection and Localization},
year = {2022},
issue_date = {Apr 2022},
publisher = {Kluwer Academic Publishers},
address = {USA},
volume = {130},
number = {4},
issn = {0920-5691},
url = {https://doi.org/10.1007/s11263-022-01578-9},
doi = {10.1007/s11263-022-01578-9},
journal = {Int. J. Comput. Vision},
month = apr,
pages = {947–969},
numpages = {23}
}

@article{hecklerkram2026mvtecad2,
  author={Heckler-Kram, Lars and Neudeck, Jan-Hendrik and Scheler, Ulla and K{\"o}nig, Rebecca and Steger, Carsten},
  title={The MVTec AD 2 Dataset: Advanced Scenarios for Unsupervised Anomaly Detection},
  journal={International Journal of Computer Vision},
  volume={134},
  pages={175},
  year={2026},
  doi={10.1007/s11263-026-02743-0}
}
\bibliographystyle{iclr2027_conference}

\appendix
\section{Appendix Overview}
In the appendix, we provide:
\begin{enumerate}[label=\Alph*.]
  \setcounter{enumi}{1}
  \item SAE training, metrics, and ablations
  \item Amplification multiplier selection
  \item Multimodal LLM ablation
  \item Anomaly detector ablation
  \item Vision backbone ablation
  \item Additional qualitative examples
  \item Example MLLM annotation prompt
\end{enumerate}

\section{SAE training, metrics, and ablations}
\label{app:sae}

\paragraph{Training and reconstruction fidelity.}
We train one SAE per category on the PatchCore residuals extracted from the
calibration images. Each SAE has width $m=64$ and TopK sparsity $k=4$, with
tied encoder and decoder weights, no biases, and unit-norm feature directions.
We optimize reconstruction MSE with Adam using a learning rate of $10^{-4}$ and
batch size 2048 for up to 1000 epochs, with early stopping based on training
reconstruction MSE (patience 50) and restoring the checkpoint with the lowest
training MSE. Table~\ref{tab:sae-fidelity} reports training and test
reconstruction MSE and cosine similarity, the number of dead features, and
recovered AUROC. Recovered AUROC is the AUROC obtained using the residual
reconstructed by the SAE without adding the reconstruction error, divided by
the original PatchCore AUROC and expressed as a percentage. The reconstructed
descriptors are rematched to the original PatchCore memory bank and rescored
with the unchanged detector. Across the 40 RQ1 categories, the SAE reconstruction recovers
95.95\% of the original PatchCore AUROC on average, while almost all features
remain active. The corresponding recovered AUROC on RobustAD is 98.74\%.

\begin{table}[h]
\caption{SAE reconstruction fidelity evaluated on the test split,
macro-averaged across categories. Recovered AUROC is reported as a percentage
of the original PatchCore AUROC.}
\label{tab:sae-fidelity}
\centering
\setlength{\tabcolsep}{3.5pt}
\scalebox{0.865}{
\begin{tabular}{l|c|c|c|c|c}
\toprule \midrule
Metric
& VisA
& MVTec AD
& MVTec LOCO
& MVTec AD 2
& All \\
& (12)
& (15)
& (5)
& (8)
& (40) \\
\midrule \midrule
Training MSE
& 21.11
& 30.41
& 26.86
& 42.86
& 29.67 \\
Test MSE
& 25.00
& 52.09
& 31.80
& 57.43
& 42.50 \\
\midrule
Training cosine
& 0.550
& 0.633
& 0.568
& 0.646
& 0.602 \\
Test cosine
& 0.504
& 0.513
& 0.550
& 0.500
& 0.512 \\
\midrule
Recovered AUROC (\%)
& 97.23
& 97.55
& 92.92
& 91.15
& 95.95 \\
\midrule
Dead features / 64
& 0.00
& 0.00
& 0.00
& 0.25
& 0.05 \\
\midrule
\bottomrule
\end{tabular}
}
\end{table}

\begin{table}[h]
\caption{SAE reconstruction fidelity evaluated on the RobustAD source test
split. Recovered AUROC is reported as a percentage of the original PatchCore
AUROC.}
\label{tab:sae-fidelity-robustad}
\centering
\setlength{\tabcolsep}{3.5pt}
\scalebox{0.865}{
\begin{tabular}{l|c|c|c|c}
\toprule \midrule
Metric
& PCB
& MetalParts
& PiledBags
& All (3) \\
\midrule \midrule
Training MSE
& 13.55
& 32.28
& 28.36
& 24.73 \\
Test MSE
& 13.49
& 32.82
& 33.10
& 26.47 \\
\midrule
Training cosine
& 0.599
& 0.553
& 0.679
& 0.610 \\
Test cosine
& 0.586
& 0.538
& 0.645
& 0.590 \\
\midrule
Recovered AUROC (\%)
& 110.37
& 84.72
& 98.74
& 98.74 \\
\midrule
Dead features / 64
& 0.00
& 0.00
& 4.00
& 1.33 \\
\midrule
\bottomrule
\end{tabular}
}
\end{table}

\paragraph{Tied and untied SAE ablation.}
We compare the tied, bias-free SAE used in the main experiments with a more
canonical untied variant that uses separate encoder and decoder weights and
learned biases, while keeping the same calibration residuals, width $m=64$,
TopK sparsity $k=4$, optimizer, and maximum number of epochs. Across the 40 RQ1
categories, the two architectures obtain similar reconstruction MSE and
recovered AUROC, while the tied SAE has substantially fewer dead features on
average, 0.05 compared with 11.43 out of 64. Even without the tying constraint,
the untied SAE still tends to learn features that contribute positively to the
PatchCore anomaly signal, with only 3.34\% of individual feature removals
increasing the raw nearest-neighbor patch distance. At comparable reconstruction
and recovered AUROC, we therefore retain the tied SAE for our intervention
setting, since individual feature removal is guaranteed not to increase this
distance, with 0\% observed violations.

\begin{table}[h]
\caption{Comparison of tied and untied SAEs across the 40 RQ1 categories.}
\label{tab:sae-architecture}
\centering
\setlength{\tabcolsep}{3.5pt}
\scalebox{0.865}{
\begin{tabular}{l|c|c}
\toprule \midrule
Metric & Tied, bias-free & Untied, with biases \\
\midrule \midrule
Test reconstruction MSE & 42.50 & 43.35 \\
Recovered AUROC (\%) & 95.95 & 95.82 \\
Dead features / 64 & 0.05 & 11.43 \\
Removals increasing patch distance (\%) & 0.00 & 3.34 \\
\midrule
\bottomrule
\end{tabular}
}
\end{table}

\paragraph{Width and sparsity sensitivity.}
We evaluate the complete pipeline for dictionary widths
$m\in\{32,64,128\}$ and TopK sparsities $k\in\{2,4,8\}$ on two categories,
VisA \emph{capsules} and MVTec AD \emph{pill}. We restrict this ablation to two
categories because there are 9 configurations per category and each requires training a new SAE, generating
new feature labels through MLLM calls, and rerunning the downstream interventions. For each of the 9 configurations, we evaluate suppression,
amplification, and joint intervention on source and corrupted data using the
same procedure as in the main experiments. Suppression and joint intervention
improve over the original detector for both categories across all tested
configurations, although the magnitude of the improvement varies significantly
and does not change monotonically with either dictionary width or TopK
sparsity. These results suggest that there may not be a single overall best configuration
among the nine tested settings, although experiments on more categories would
be needed to establish this more clearly.

\begin{table}[h]
\caption{Width and TopK sensitivity on VisA \emph{capsules} and MVTec AD
\emph{pill}, macro-averaged across the two categories.}
\label{tab:sae-width-topk}
\centering
\setlength{\tabcolsep}{3.5pt}
\scalebox{0.865}{
\begin{tabular}{c|c|c|c|c|c|c|c}
\toprule \midrule
Width & TopK
& \multicolumn{3}{c|}{Source}
& \multicolumn{3}{c}{Corrupted} \\
$m$ & $k$
& Suppression
& Amplification
& Joint
& Suppression
& Amplification
& Joint \\
\midrule \midrule
\multicolumn{2}{l|}{Original PatchCore}
& \multicolumn{3}{c|}{0.8592}
& \multicolumn{3}{c}{0.7836} \\
\midrule \midrule
32  & 2 & 0.8816 & \textbf{0.9418} & \textbf{0.9465}
        & 0.7998 & 0.8544 & 0.8571 \\
32  & 4 & 0.8924 & 0.9263 & 0.9373
        & 0.8027 & 0.8545 & 0.8543 \\
32  & 8 & 0.8903 & 0.9123 & 0.8903
        & 0.8023 & 0.8414 & 0.8023 \\
\midrule
64  & 2 & 0.8858 & 0.8821 & 0.8965
        & 0.7969 & 0.8097 & 0.8126 \\
\textbf{64} & \textbf{4}
        & 0.8890 & 0.9245 & 0.9265
        & 0.7951 & \textbf{0.8545} & \textbf{0.8580} \\
64  & 8 & 0.8975 & 0.8834 & 0.8975
        & 0.8021 & 0.8001 & 0.8021 \\
\midrule
128 & 2 & 0.8778 & 0.8830 & 0.8876
        & 0.7954 & 0.8037 & 0.7990 \\
128 & 4 & 0.8846 & 0.8863 & 0.8739
        & 0.7990 & 0.7964 & 0.7885 \\
128 & 8 & \textbf{0.9071} & 0.9116 & 0.9071
        & \textbf{0.8067} & 0.8164 & 0.8067 \\
\midrule
\bottomrule
\end{tabular}
}
\end{table}

\paragraph{Dependence on reconstruction error.}
Our main intervention preserves the SAE reconstruction error
$\boldsymbol{\epsilon}_p=\mathbf{r}_p-\hat{\mathbf{r}}_p$ by design.
Because the SAE is trained only on calibration residuals, replacing the
original residual with its sparse reconstruction would discard information not
captured by the dictionary learned from the calibration set. This would be
undesirable in anomaly detection, where held-out anomalies need not resemble
those available during calibration. Preserving $\boldsymbol{\epsilon}_p$
therefore keeps the residual information outside the SAE representation while
allowing the intervention to edit only the decoded sparse contributions.

We nevertheless omit $\boldsymbol{\epsilon}_p$ as a diagnostic to measure how
the SAE reconstruction performs on its own and whether the intervention gains
remain when evaluation is restricted to this representation. Amplification and
joint intervention select their multiplier independently with and without
reconstruction error using the same clean-calibration procedure as in the main
experiments.

\begin{table}[h]
\caption{Effect of retaining the SAE reconstruction error across the 40 RQ1
categories, macro-averaged across categories. For no intervention, the
error-retained condition corresponds to original PatchCore, while the
error-omitted condition corresponds to the SAE-only reconstruction.}
\label{tab:sae-reconstruction-error}
\centering
\setlength{\tabcolsep}{3.5pt}
\scalebox{0.865}{
\begin{tabular}{l|c|c|c|c}
\toprule \midrule
& \multicolumn{2}{c|}{Source}
& \multicolumn{2}{c}{Corrupted} \\
Intervention
& Error retained
& Error omitted
& Error retained
& Error omitted \\
\midrule \midrule
None
& 0.8724
& 0.8368
& 0.8066
& 0.7760 \\
Suppression
& 0.8808
& 0.8516
& 0.8126
& 0.8102 \\
Amplification
& 0.8819
& 0.8524
& 0.8152
& 0.7964 \\
Joint
& \textbf{0.8857}
& \textbf{0.8551}
& \textbf{0.8210}
& \textbf{0.8136} \\
\midrule
\bottomrule
\end{tabular}
}
\end{table}

Without reconstruction error, the unedited SAE representation reaches 0.8368
source AUROC, substantially below the original PatchCore value of 0.8724.
A complementary source-only diagnostic shows the opposite component: retaining
only the reconstruction error yields 0.8697 AUROC, close to the original
detector. These component evaluations are not an additive decomposition of
AUROC because memory-bank rematching and image-level scoring are nonlinear.

Importantly, the intervention gains remain visible when evaluation is
restricted to the SAE reconstruction. Relative to the no-intervention SAE-only
baseline, suppression, amplification, and joint intervention increase source
AUROC from 0.8368 to 0.8516, 0.8524, and 0.8551, respectively. Under corruption,
they increase AUROC from 0.7760 to 0.8102, 0.7964, and 0.8136. Thus, the
reconstruction error remains important for preserving the detector information
outside the sparse representation, while the benefits of the interventions are
also present within the SAE representation itself.

We additionally test random feature interventions without reconstruction error
on source and corrupted data. The results are shown in
Table~\ref{tab:sae-no-error-random}.

\begin{table}[h]
\caption{AUROC for random interventions without SAE reconstruction error
across the 40 RQ1 categories, macro-averaged across categories. None
corresponds to the unedited SAE-only reconstruction. Random interventions use
a single feature assignment.}
\label{tab:sae-no-error-random}
\centering
\begin{tabular}{l|c|c}
\toprule \midrule
Intervention & Source & Corrupted \\
\midrule \midrule
None & \textbf{0.8368} & \textbf{0.7760} \\
Suppression & 0.7807 & 0.7274 \\
Amplification & 0.8273 & 0.7715 \\
Joint & 0.7899 & 0.7348 \\
\midrule
\bottomrule
\end{tabular}
\end{table}

Random interventions reduce performance relative to the unedited SAE-only
representation, but do not collapse the detector even when the reconstruction
error is absent. The same pattern holds under corruptions. On source data,
AutoInterp remains higher under all three interventions, showing that its gains
within the SAE representation depend on which features are edited rather than
on sparse intervention alone.

\section{Amplification multiplier selection}
\label{app:multiplier-sensitivity}

We evaluate how the amplification multiplier affects performance for both
amplification-only and joint intervention. We vary
$\alpha\in\{1,1.5,2,3,4,5\}$ across the 40 RQ1 categories. In the
amplification-only setting, anomaly features are multiplied by $\alpha$, while
distractor and uncertain features are left unchanged. In the joint setting,
distractor features are additionally suppressed. Thus, $\alpha=1$ leaves the
original detector unchanged for amplification-only and reduces the joint
intervention to distractor suppression. We also report the setting used in the
main experiments, where $\alpha$ is selected independently for each category
using clean calibration AUROC and then kept fixed during source and corrupted
evaluation.

\begin{table}[h]
\caption{Effect of the anomaly-feature amplification multiplier on
amplification-only and joint intervention across the 40 RQ1 categories,
macro-averaged across categories. Best results in each column are in bold.}
\label{tab:amplification-multiplier}
\centering
\setlength{\tabcolsep}{3.5pt}
\scalebox{0.865}{
\begin{tabular}{l|c|c|c|c}
\toprule \midrule
& \multicolumn{2}{c|}{Source}
& \multicolumn{2}{c}{Corrupted} \\
Multiplier
& Amplification
& Joint
& Amplification
& Joint \\
\midrule \midrule
$1\times$
& 0.8724
& 0.8808
& 0.8066
& 0.8126 \\
$1.5\times$
& 0.8761
& 0.8831
& 0.8155
& 0.8199 \\
$2\times$
& 0.8785
& 0.8836
& 0.8210
& 0.8245 \\
$3\times$
& 0.8801
& 0.8789
& \textbf{0.8246}
& \textbf{0.8259} \\
$4\times$
& 0.8745
& 0.8735
& 0.8236
& 0.8237 \\
$5\times$
& 0.8703
& 0.8702
& 0.8210
& 0.8212 \\
\midrule
Calibration-selected
& \textbf{0.8819}
& \textbf{0.8857}
& 0.8152
& 0.8210 \\
\midrule
\bottomrule
\end{tabular}
}
\end{table}

For joint intervention, performance increases for moderate amplification
strengths and decreases on source data for larger multipliers. Among the fixed
values, $2\times$ obtains the highest source AUROC and $3\times$ the highest
corrupted AUROC. Amplification-only shows a similar sensitivity to the
multiplier, with performance increasing up to $3\times$ on both source and
corrupted data before decreasing at larger values. The calibration-selected
setting obtains the highest source AUROC for both interventions, but does not
maximize corrupted AUROC.

These results also highlight a limitation of the current multiplier selection
strategy. Selecting $\alpha$ independently on a small clean calibration set may
overfit category-specific variation, while using a single fixed multiplier
across all categories does not account for differences in intervention
sensitivity. More robust strategies for selecting intervention strength may
therefore be worth exploring.

\section{Multimodal LLM ablation}
\label{app:mllm-ablation}
We evaluate the sensitivity of the method to the MLLM used for feature
interpretation on a fixed subset of ten categories. We compare GPT-5.6 Sol,
used in the main experiments, with the open-weight Gemma 4 31B-IT and
Qwen3.5-27B. The detector, SAE, interpretation protocol, and downstream
intervention procedure are kept unchanged. For each annotator, we evaluate
suppression, amplification-only, and joint intervention on source and corrupted
data.

\begin{table}[h]
\caption{Effect of the MLLM used for feature interpretation on a fixed subset
of ten categories, macro-averaged across categories.}
\label{tab:mllm-ablation}
\centering
\setlength{\tabcolsep}{3.5pt}
\scalebox{0.865}{
\begin{tabular}{l|c|c|c|c|c|c}
\toprule \midrule
& \multicolumn{3}{c|}{Source}
& \multicolumn{3}{c}{Corrupted} \\
MLLM
& Suppression
& Amplification
& Joint
& Suppression
& Amplification
& Joint \\
\midrule \midrule
Original PatchCore
& \multicolumn{3}{c|}{0.8653}
& \multicolumn{3}{c}{0.7951} \\
\midrule \midrule
GPT-5.6 Sol
& 0.8779
& 0.8808
& 0.8849
& 0.8024
& 0.8157
& \textbf{0.8243} \\
Gemma 4 31B-IT
& \textbf{0.8786}
& \textbf{0.8853}
& \textbf{0.8899}
& 0.8006
& \textbf{0.8165}
& 0.8207 \\
Qwen3.5-27B
& 0.8773
& 0.8708
& 0.8791
& \textbf{0.8043}
& 0.8055
& 0.8149 \\
\midrule
\bottomrule
\end{tabular}
}
\end{table}

Joint intervention improves over the original detector on both source and
corrupted data for all three MLLMs. The exact semantic labels are more
model-dependent, with pairwise agreement of 81.1\% between GPT and Gemma,
69.5\% between GPT and Qwen, and 67.8\% between Gemma and Qwen. Despite these
differences, the downstream intervention remains beneficial across all three
annotators, suggesting that the observed effect is not specific to the MLLM
used in the main experiments.

\section{Anomaly detector ablation}
\label{app:detector-ablation}

To test whether our approach extends beyond nearest-neighbor anomaly detection,
we repeat the pipeline with FRE \citep{fre}, a feature-reconstruction detector,
on five categories: Capsules, Pill, Grid, Transistor, and Pushpins. We use the
native Anomalib FRE implementation with a frozen DINOv3 ViT-L/16 backbone,
extracting representations from \texttt{blocks.23} at an input resolution of
$256\times256$. FRE applies $2\times2$ average pooling and flattens the resulting
$1024\times8\times8$ feature tensor before reconstructing it with a tied linear
autoencoder with a 220-dimensional latent representation. The FRE autoencoder
is trained with mean squared reconstruction error and Adam with learning rate
$10^{-3}$, batch size 128, and a fixed 440 epochs.

The resulting FRE reconstruction residual is decomposed using a separate tied
TopK SAE with 64 features and $k=4$ for each category. Automated
interpretability and interventions follow the same protocol as in the main
experiments. Distractor features are suppressed, anomaly features are amplified,
and uncertain features are left unchanged. Amplification-only and joint
intervention select their multiplier from $\{1,1.5,2,3,4,5\}$ using source
calibration AUROC and keep it fixed under corruption. PatchCore uses its main
configuration on the same five categories.

\begin{table}[h]
\caption{Anomaly detector ablation on five categories. Deltas are AUROC
percentage points relative to the corresponding original detector. Best
intervention results for each detector and evaluation setting are in bold.}
\label{tab:detector-ablation}
\centering
\setlength{\tabcolsep}{3.5pt}
\scalebox{0.865}{
\begin{tabular}{l|l|c|c|c|c}
\toprule \midrule
Detector & Intervention
& \multicolumn{2}{c|}{Source}
& \multicolumn{2}{c}{Corrupted} \\
& & AUROC & $\Delta$ & AUROC & $\Delta$ \\
\midrule \midrule
\multirow{4}{*}{PatchCore}
& Original
& 0.8969 & -- & 0.8428 & -- \\
\cmidrule(l){2-6}
& Suppression
& 0.9105 & +1.36 & 0.8464 & +0.37 \\
& Amplification
& 0.9166 & +1.96 & 0.8714 & +2.86 \\
& Joint
& \textbf{0.9255} & \textbf{+2.86}
& \textbf{0.8716} & \textbf{+2.88} \\
\midrule \midrule
\multirow{4}{*}{FRE}
& Original
& 0.8475 & -- & 0.7488 & -- \\
\cmidrule(l){2-6}
& Suppression
& \textbf{0.8679} & \textbf{+2.03}
& 0.7462 & -0.26 \\
& Amplification
& 0.8501 & +0.26
& \textbf{0.7602} & \textbf{+1.14} \\
& Joint
& 0.8675 & +2.00
& 0.7479 & -0.09 \\
\midrule
\bottomrule
\end{tabular}
}
\end{table}

With FRE, suppression and joint intervention improve source AUROC by 2.03 and
2.00 points, respectively. Under synthetic corruptions, however, their average
effects are close to zero, with changes of -0.26 and -0.09 points. Amplification
shows the opposite pattern, producing only a small source gain of 0.26 points
but improving corrupted AUROC by 1.14 points. The corresponding PatchCore
interventions produce positive average gains in both evaluations on the same
five categories. These results provide some preliminary evidence that the
intervention framework could extend to a feature-reconstruction detector.
However, experiments on more categories would be needed to draw stronger
conclusions.

\section{Vision backbone ablation}
\label{app:backbone-ablation}

To test whether the effect of our interventions transfers beyond the DINOv3
backbone used in the main experiments, we repeat the complete pipeline with
SwinV2-L and WideResNet-50-2 on the same five categories used in the detector
ablation, Capsules, Pill, Grid, Transistor, and Pushpins. SwinV2-L uses
ImageNet-22K pretrained weights fine-tuned on ImageNet-1K, while
WideResNet-50-2 uses ImageNet-1K pretrained weights. DINOv3-L/16 uses the
main-paper representations from \texttt{blocks.11} and \texttt{blocks.23}.
For WideResNet-50-2, we use \texttt{layer2} and \texttt{layer3}, following the
default PatchCore configuration in Anomalib. For SwinV2-L, we use  \texttt{layers.1} and \texttt{layers.2}. The PatchCore
coreset ratio, SAE width of 64 with TopK $k=4$, annotation protocol,
calibration split, intervention procedure, and corruption methodology are kept fixed.

\begin{table}[h]
\caption{Vision backbone ablation over five categories. Deltas are AUROC
percentage points relative to the corresponding original detector. Largest
intervention gains are in bold.}
\label{tab:backbone-ablation}
\centering
\setlength{\tabcolsep}{3.5pt}
\scalebox{0.865}{
\begin{tabular}{l|c|c|c|c|c|c}
\toprule \midrule
& \multicolumn{3}{c|}{Source}
& \multicolumn{3}{c}{Corrupted} \\
Backbone
& Original
& Suppression $\Delta$
& Joint $\Delta$
& Original
& Suppression $\Delta$
& Joint $\Delta$ \\
\midrule \midrule
DINOv3-L/16
& 0.8969
& +1.36
& \textbf{+2.86}
& 0.8428
& +0.37
& \textbf{+2.88} \\
SwinV2-L
& 0.8073
& \textbf{+1.41}
& +1.78
& 0.7177
& \textbf{+1.02}
& +1.22 \\
WideResNet-50-2
& 0.9213
& -0.19
& -0.19
& 0.8297
& -0.07
& -0.07 \\
\midrule
\bottomrule
\end{tabular}
}
\end{table}

SwinV2 has a lower original AUROC than the other backbones, driven in part by
Pushpins, where PatchCore performs at approximately chance level in both source
and corrupted evaluation. On the same category, DINOv3-L/16 reaches 0.7666 and
0.6994 AUROC on source and corrupted data, respectively, while
WideResNet-50-2 reaches 0.7286 and 0.6807.

The positive average effect obtained with DINOv3 also transfers to SwinV2 on
these five categories. Joint intervention improves source and corrupted AUROC
by 2.86 and 2.88 points with DINOv3, and by 1.78 and 1.22 points with SwinV2.
In contrast, the average effect with WideResNet-50-2 is close to zero, with
changes of -0.19 and -0.07 points.


\section{Additional qualitative examples}
\label{app:qualitative}


\newcommand{\qualimgcell}[1]{%
  \begin{minipage}[t]{0.235\linewidth}\centering
    \includegraphics[width=\linewidth]{#1}
  \end{minipage}}

\newcommand{\qualfeaturepanel}[5]{%
\begin{minipage}[t]{0.487\textwidth}
\centering

\parbox[t][11.2mm][t]{\linewidth}{\centering
{\small (#1) Feature labeled as #2}\\[-0.35mm]
{\scriptsize #3}\\[-0.25mm]
{\scriptsize\itshape ``#4''}}

\begin{minipage}[t]{0.235\linewidth}\centering
{\scriptsize Test image}
\end{minipage}\hfill
\begin{minipage}[t]{0.745\linewidth}\centering
{\scriptsize Top-activating calibration examples}
\end{minipage}\\[0.45mm]

\makebox[\linewidth][c]{%
\qualimgcell{#5/held_out}\hfill
\qualimgcell{#5/annotation_heatmap_01.png}\hfill
\qualimgcell{#5/annotation_heatmap_02.png}\hfill
\qualimgcell{#5/annotation_heatmap_03.png}}\\[1.05mm]

\begin{minipage}[t]{0.235\linewidth}\centering
{\scriptsize\mbox{SAE feature map}}
\end{minipage}\hfill
\begin{minipage}[t]{0.745\linewidth}\centering
{\scriptsize PatchCore anomaly maps}
\end{minipage}\\[0.45mm]

\makebox[\linewidth][c]{%
\qualimgcell{#5/held_out_unit_heatmap.png}\hfill
\qualimgcell{#5/patchcore_before_ablation.png}\hfill
\qualimgcell{#5/patchcore_after_ablation.png}\hfill
\qualimgcell{#5/patchcore_after_unit_amplification.png}}\\[-0.2mm]

\begin{minipage}[t]{0.235\linewidth}\centering~\end{minipage}\hfill
\begin{minipage}[t]{0.235\linewidth}\centering
{\scriptsize Original}
\end{minipage}\hfill
\begin{minipage}[t]{0.235\linewidth}\centering
{\scriptsize Suppressed}
\end{minipage}\hfill
\begin{minipage}[t]{0.235\linewidth}\centering
{\scriptsize Amplified}
\end{minipage}

\end{minipage}}

Here we provide additional examples of SAE features and their intervention effects.
Italic text gives the MLLM-generated feature description. For each feature, the
top row shows a test image and three top-activating calibration examples.
The bottom row shows the test SAE feature map and PatchCore anomaly maps
before intervention, after suppression, and after $2\times$ amplification.
The three PatchCore anomaly maps for each example use a shared min-max color
scale.

\vspace{1.5mm}

\qualfeaturepanel
{a}
{anomaly}
{VisA PCB1, SAE feature 3}
{bent or angled connector pin}
{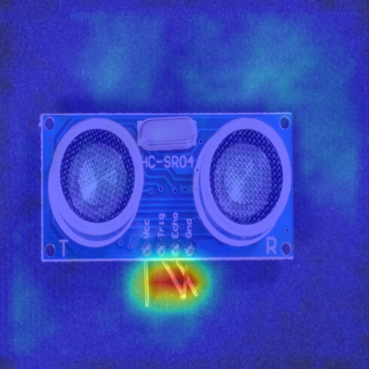}
\hfill
\qualfeaturepanel
{b}
{distractor}
{MVTec AD Screw, SAE feature 15}
{faint isolated dark speck on the background}
{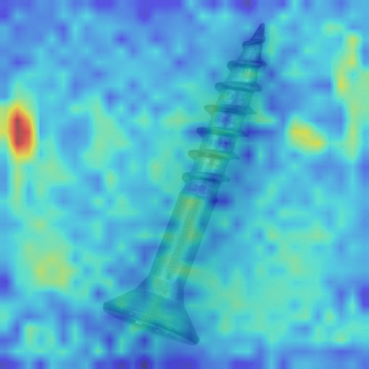}

\par\vspace{3.2mm}

\qualfeaturepanel
{c}
{anomaly}
{VisA Chewing Gum, SAE feature 35}
{torn, gouged chewing-gum surface with exposed brown interior}
{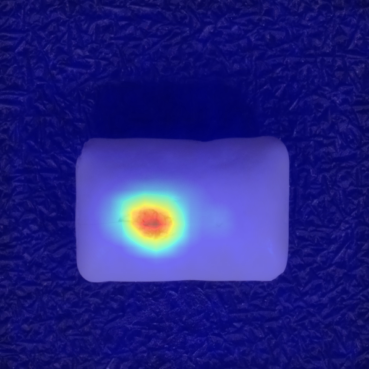}
\hfill
\qualfeaturepanel
{d}
{distractor}
{VisA Cashew, SAE feature 7}
{glossy, strongly highlighted background fibers}
{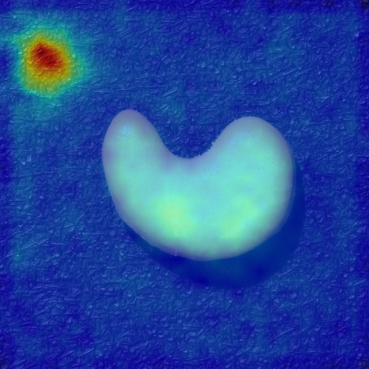}

\par\vspace{3.2mm}

\qualfeaturepanel
{e}
{anomaly}
{VisA PCB2, SAE feature 44}
{excess solder blobs or solder bridging on PCB components}
{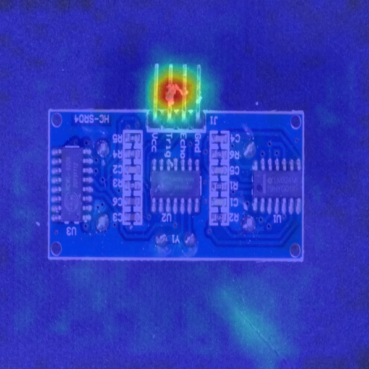}
\hfill
\qualfeaturepanel
{f}
{anomaly}
{MVTec AD Cable, SAE feature 8}
{elongated protruding copper strands}
{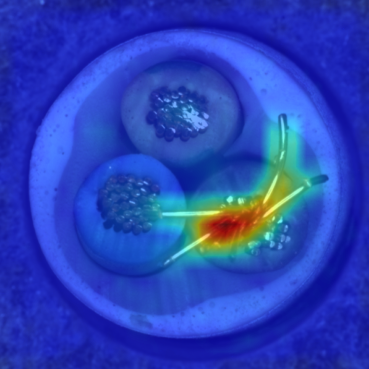}

\par\vspace{3.2mm}

\qualfeaturepanel
{g}
{anomaly}
{MVTec AD Zipper, SAE feature 25}
{tangled protruding threads or fibers}
{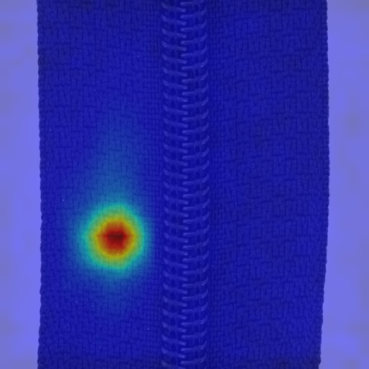}
\hfill
\qualfeaturepanel
{h}
{uncertain}
{MVTec AD Capsule, SAE feature 48}
{capsule surface defects or markings}
{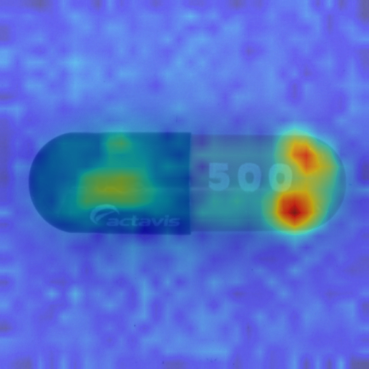}

\par\vspace{3.2mm}

\qualfeaturepanel
{i}
{anomaly}
{VisA PCB1, SAE feature 16}
{{\fontsize{6.1pt}{6.7pt}\selectfont
missing or severely damaged transducer mesh exposing a smooth inner disc}}
{figures/qualitative/pcb1_feature16}
\hfill
\qualfeaturepanel
{j}
{distractor}
{MVTec AD 2 Walnuts, SAE feature 26}
{dark background with coarse diagonal ridges}
{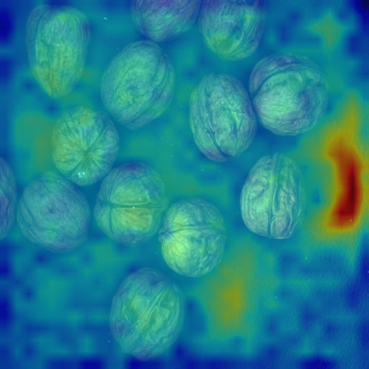}

\par\vspace{3.2mm}

\qualfeaturepanel
{k}
{anomaly}
{VisA PCB2, SAE feature 50}
{bent or misaligned header pins}
{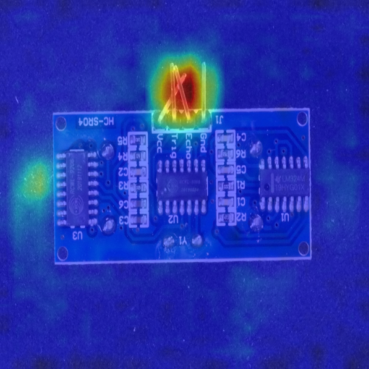}
\hfill
\qualfeaturepanel
{l}
{anomaly}
{MVTec AD Transistor, SAE feature 61}
{transistor lead tip not inserted into a board hole}
{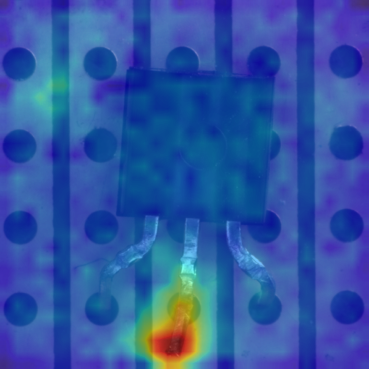}

\par\vspace{3.2mm}

\qualfeaturepanel
{m}
{anomaly}
{VisA PCB2, SAE feature 9}
{missing or severely displaced connector pins/exposed header pads}
{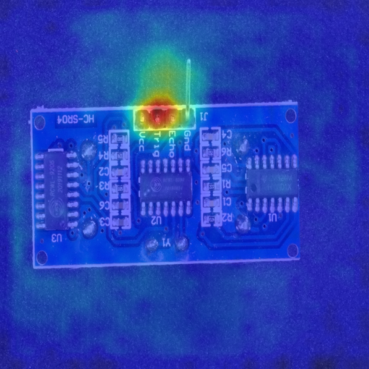}
\hfill
\qualfeaturepanel
{n}
{distractor}
{MVTec AD Toothbrush, SAE feature 58}
{thin brown foreign filament or scratch on the dark background}
{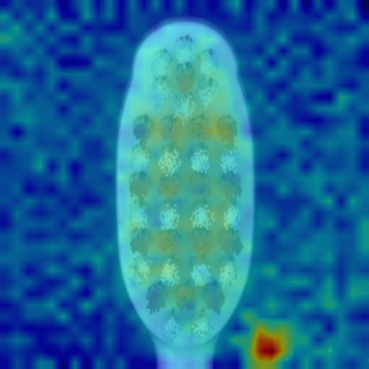}

\par\vspace{3.2mm}

\qualfeaturepanel
{o}
{anomaly}
{VisA PCB3, SAE feature 44}
{excess solder blobs or solder bridging on PCB components}
{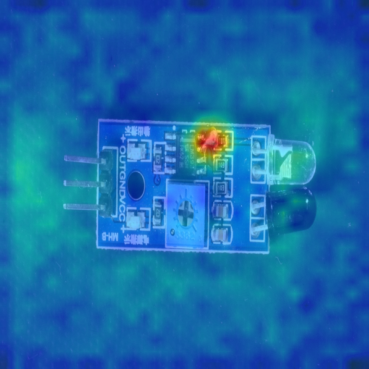}
\hfill
\qualfeaturepanel
{p}
{distractor}
{VisA Capsules, SAE feature 48}
{dark smudge or stain on the background fabric}
{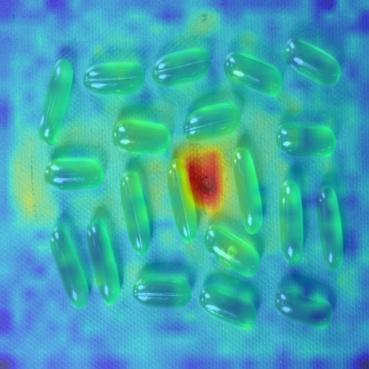}

\par\vspace{3.2mm}

\qualfeaturepanel
{q}
{anomaly}
{VisA PCB4, SAE feature 24}
{missing or severely damaged USB connector shell}
{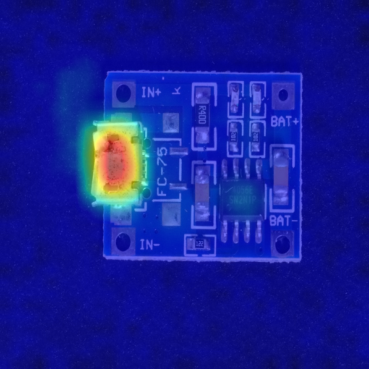}
\hfill
\qualfeaturepanel
{r}
{anomaly}
{MVTec AD Hazelnut, SAE feature 10}
{exposed pale inner shell along deep cracks or holes}
{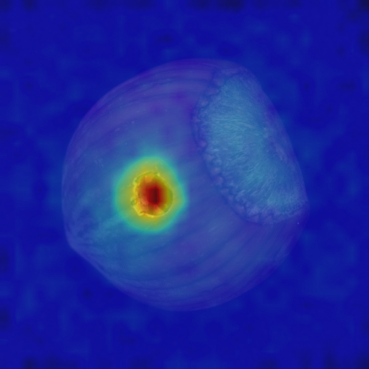}

\section{Example MLLM annotation prompt}
\label{app:autointerp-prompt}

\newenvironment{AutoInterpPromptBlock}[1]{%
  \par\medskip
  \noindent\textbf{#1}\par\smallskip
  \begingroup
  \small
  \setlength{\parindent}{0pt}%
  \setlength{\parskip}{0.55em}%
}{%
  \par\endgroup\medskip
}

\newcommand{\AutoInterpPromptImage}[3][0.78\linewidth]{%
  \par\smallskip
  \noindent\begin{minipage}{\linewidth}
    \centering
    {\footnotesize\textbf{#2}\par}\smallskip
    \includegraphics[width=#1]{#3}
  \end{minipage}
  \par\smallskip
}

\begin{AutoInterpPromptBlock}{System prompt}
You are an expert in visual anomaly detection for industrial inspection. The
dataset is RobustAD, category PCB. Official benchmark normality specification
for this category: The PCB sub-dataset captures the challenges of finding
subtle scratches, soldering melts, and missing parts which comprise of the most
common defects encountered during inspection of Printed Circuit Boards in
electronics and semiconductor manufacturing. Defects: Missing components, scratches, soldering melts
\end{AutoInterpPromptBlock}

\begin{AutoInterpPromptBlock}{User prompt}
We're studying a neuron in a sparse autoencoder used to interpret a PatchCore anomaly detector.

You will first be shown a reference grid of known-normal images.

You will then be shown high-activation context panels. Each panel contains the full image with the selected PatchCore region boxed, the neuron's spatial unit activation map, and a magnified crop with the same region boxed. All displayed activation maps use the same unit-specific color scale.

The reported activation is read out at the red-boxed PatchCore location; it does not describe the whole image. A zero-activation comparison means that the unit is inactive at the selected comparison location, even if another part of the image contains an anomaly or activates the unit. Compare the boxed regions, using the full images only as context.

The colors in the unit activation map and the red boxes are visual aids; they are not present in the source image. Use the map only to locate activation, and infer visual content from the full image and high-activation region.

Treat the activation map and box as approximate spatial cues, since transformer tokens mix contextual information.

The final group contains zero-activation comparison panels. Each panel shows the complete source image with the selected PatchCore region boxed and the same region magnified alongside it.

Your task is to identify the common visual concept in the selected high-activation regions that is absent from the selected zero-activation comparison regions.

\AutoInterpPromptImage[0.48\linewidth]
  {Known-normal reference grid}
  {figures/prompt/normal\_reference\_grid.png}

\AutoInterpPromptImage[0.78\linewidth]
  {High-activation context panel 1, selected-location activation: 411.15}
  {figures/prompt/high\_activation\_1.png}
\AutoInterpPromptImage[0.78\linewidth]
  {High-activation context panel 2, selected-location activation: 392.50}
  {figures/prompt/high\_activation\_2.png}
\AutoInterpPromptImage[0.78\linewidth]
  {High-activation context panel 3, selected-location activation: 378.06}
  {figures/prompt/high\_activation\_3.png}
\AutoInterpPromptImage[0.78\linewidth]
  {High-activation context panel 4, selected-location activation: 321.88}
  {figures/prompt/high\_activation\_4.png}
\AutoInterpPromptImage[0.78\linewidth]
  {High-activation context panel 5, selected-location activation: 278.46}
  {figures/prompt/high\_activation\_5.png}

\AutoInterpPromptImage[0.52\linewidth]
  {Zero-activation comparison panel 1}
  {figures/prompt/zero\_activation\_1.png}
\AutoInterpPromptImage[0.52\linewidth]
  {Zero-activation comparison panel 2}
  {figures/prompt/zero\_activation\_2.png}
\AutoInterpPromptImage[0.52\linewidth]
  {Zero-activation comparison panel 3}
  {figures/prompt/zero\_activation\_3.png}
\AutoInterpPromptImage[0.52\linewidth]
  {Zero-activation comparison panel 4}
  {figures/prompt/zero\_activation\_4.png}
\AutoInterpPromptImage[0.52\linewidth]
  {Zero-activation comparison panel 5}
  {figures/prompt/zero\_activation\_5.png}

Now that you've analyzed the patches, provide just a concise concept name and its semantic type.

Name the visual attribute or combination of attributes that clearly appears to cause the activation, rather than describing the full patch or including weak incidental correlations. Avoid vague names like `distinctive pattern'.

Classify the concept by its relevance to the inspection target, not by rarity alone. Use anomaly when it indicates an abnormality of the target or its expected configuration. Use distractor when it captures something unrelated to the target or an ordinary target variation. Use uncertain when the evidence is mixed or unclear.

Identify the localized visual feature that distinguishes the activating examples. Use the normal reference grid only to understand normal appearance, and do not classify from activation magnitude alone. Never name or classify colors, textures, or boundaries introduced by the unit activation map visualization.

Return a JSON object with exactly two keys: concept\_name and semantic\_type.
\end{AutoInterpPromptBlock}

\end{document}